\documentclass{article}

\usepackage{arxiv}

\usepackage[utf8]{inputenc} 
\usepackage[T1]{fontenc}    
\usepackage{hyperref}       
\usepackage{url}            
\usepackage{booktabs}       
\usepackage{amsfonts}       
\usepackage{nicefrac}       
\usepackage{microtype}      
\usepackage{graphicx}
\usepackage{amsmath}
\usepackage{multirow}
\usepackage{xcolor}
\usepackage{subcaption}
\usepackage[section]{placeins}  

\graphicspath{{./figures/}}

\title{Relax: An Asynchronous Reinforcement Learning Engine \\ for Omni-Modal Post-Training at Scale}

\author{
\textbf{Liujie Zhang}$^1$ \quad
\textbf{Benzhe Ning}$^1$ \quad
\textbf{Rui Yang}$^1$ \quad
\textbf{Xiaoyan Yu}$^1$ \quad
\textbf{Jiaxing Li}$^1$ \quad
\textbf{Lumeng Wu}$^2$ \quad
\textbf{Jia Liu}$^3$ \quad \\
\textbf{Minghao Li}$^1$ \quad
\textbf{Weihang Chen}$^1$ \quad
\textbf{Weiqi Hu}$^1$ \quad
\textbf{Lei Zhang}$^1$ \quad
\\
\\
$^1$AI Platform, Xiaohongshu Inc \\
$^2$The University of Hong Kong \quad $^3$University of Science and Technology of China
}

\begin{document}
\maketitle

\begin{abstract}
Reinforcement learning (RL) post-training has proven effective at
unlocking reasoning, self-reflection, and tool-use capabilities in
large language models.
As models extend to omni-modal inputs and agentic multi-turn
workflows, RL training systems face three interdependent challenges:
heterogeneous data flows, operational robustness at scale, and the
staleness--throughput tradeoff.
We present \textbf{Relax} (Reinforcement Engine Leveraging Agentic
X-modality), an open-source RL training engine that addresses these
challenges through three co-designed architectural layers.
First, an \emph{omni-native architecture} builds multimodal support
into the full stack---from data preprocessing and modality-aware
parallelism to inference generation---rather than retrofitting it
onto a text-centric pipeline.
Second, each RL role runs as an independent, fault-isolated service
that can be scaled, recovered, and upgraded without global
coordination.
Third, service-level decoupling enables asynchronous training via the
TransferQueue data bus, where a single staleness parameter smoothly
interpolates among on-policy, near-on-policy, and fully asynchronous
execution.
Relax achieves a 1.20$\times$ end-to-end speedup over veRL on
Qwen3-4B on-policy training.
Its fully async mode delivers a 1.76$\times$ speedup over colocate
on Qwen3-4B and a 2.00$\times$ speedup on Qwen3-Omni-30B, while
all modes converge to the same reward level.
Relax supports R3 (Rollout Routing Replay)~\cite{ma2025r3} for MoE models with only 1.9\%
overhead, compared to 32\% degradation in veRL under the same
configuration.
It further demonstrates stable omni-modal RL convergence on
Qwen3-Omni across image, text, and audio, sustaining over 2{,}000
steps on video without degradation.
Relax is available at \url{https://github.com/redai-infra/Relax}.
\end{abstract}

\section{Introduction}
\label{sec:introduction}

\subsection{From Text-Only to Omni-Modal and Agentic RL}
\label{subsec:crossroads}

Reinforcement learning has become a critical post-training technique for improving reasoning in large language models.
Landmark systems such as DeepSeek-R1~\cite{deepseek2025r1} and the OpenAI o1/o3 series have demonstrated that RL can unlock capabilities that supervised fine-tuning alone cannot produce, including self-reflection, step-by-step verification, and complex multi-hop reasoning.
In parallel, the algorithmic toolkit has expanded rapidly: from PPO~\cite{schulman2017ppo} and its memory-efficient variant GRPO~\cite{shao2024deepseekmath}, to the decoupled clip and dynamic sampling strategies of DAPO~\cite{yu2025dapo}, and further to reward-model-free paradigms such as RLVR, RL post-training has matured from research exploration to production deployment.
This momentum has spurred a wave of open-source RL training frameworks, including veRL~\cite{sheng2024hybridflow}, OpenRLHF~\cite{hu2024openrlhf}, AReaL~\cite{fu2025areal}, AsyncFlow~\cite{han2025asyncflow}, ROLL~\cite{wang2025roll}, ProRL~\cite{zhang2026prorl}, and Slime~\cite{slime2025}, each advancing the state of the art along complementary dimensions such as hybrid control flow, asynchronous scheduling, and multi-turn agentic rollouts.

While algorithmic and systems progress continues, the \emph{training paradigm itself} is undergoing a fundamental shift along two axes.
First, models are becoming \textbf{omni-modal}: from vision-language models such as Qwen3-VL to unified omni-modal architectures such as Qwen3-Omni and Qwen3.5-Omni, models increasingly consume images, video, and audio alongside text, and their RL training must handle all of these modalities from the data pipeline through the parallel execution strategy to the inference backend.
Second, training is becoming \textbf{agentic}: rather than single-turn prompt--response pairs, RL workflows now involve multi-turn reasoning, tool calling, and search-augmented generation, where the model interacts with external environments across many steps before a reward signal is produced.
These two trends interact: omni-modal data makes workloads more heterogeneous and failure-prone, while agentic rollouts introduce variable-length, multi-step trajectories. Addressing them effectively requires rethinking the training system as a whole rather than patching individual components.

\subsection{Three Fundamental Challenges}
\label{subsec:challenges}

We identify three fundamental challenges that jointly define the design space for next-generation RL training systems.
Rather than being independent concerns, they form a causal progression from data heterogeneity through operational robustness to execution flexibility, each shaping the requirements of the next.

\paragraph{Heterogeneous modality pipelines.}
When RL training extends from text to images, video, and audio, every layer of the data pipeline must be redesigned.
A single training batch may contain image patches of varying resolutions, variable-length video frame sequences, and audio waveforms sampled at different rates, all interspersed with text tokens.
These modalities differ not only in representation and size, where a high-resolution image can be 50$\times$ the token count of its accompanying text prompt, but also in preprocessing latency, memory footprint, and the parallel strategy required for efficient encoding (e.g., ViT sharding for images vs.\ streaming decoding for audio).
Text-first frameworks that bolt on multimodal support via ad hoc preprocessing wrappers and special-cased code paths struggle to generalize across modalities and often introduce fragile, hard-to-maintain pipelines.

\paragraph{Operational robustness at scale.}
The heterogeneity introduced by omni-modal data directly amplifies operational challenges.
Omni-modal workloads exhibit more severe long-tail latency distributions, where a batch containing a 30-second video clip may take 10$\times$ longer to process than a text-only batch, and are more prone to out-of-memory failures when large visual inputs coexist with long-context language generation.
At the scale of hundreds to thousands of GPUs running for days, training must tolerate hardware failures, NCCL timeouts, and individual service crashes without restarting the entire job.
This requires minute-level fault recovery, elastic scaling of individual roles, and per-role lifecycle management, which go well beyond traditional checkpoint-and-restart strategies.

\paragraph{Execution decoupling and policy flexibility.}
Service decoupling opens a further opportunity: once RL roles are independently deployable, the execution schedule no longer needs to be globally synchronous.
In a typical RL training step, data flows through rollout generation, reward computation, advantage estimation, and policy gradient training.
If these stages execute synchronously, the trainer GPU sits idle while waiting for the slowest rollout, a problem exacerbated by omni-modal long-tail latency.
Fully asynchronous execution can recover this idle time, but introduces the question of data staleness: training on rollouts generated by an older policy version may harm convergence.
A practical system must therefore provide a unified abstraction for smoothly switching between on-policy, near-on-policy, and off-policy training modes, allowing practitioners to navigate the throughput--staleness tradeoff with a single configuration knob rather than wholesale code changes.

\subsection{Contributions}
\label{subsec:contributions}

The central insight of this work is that these three challenges are tightly coupled and most effectively addressed through \emph{co-design}.
They are linked by directed dependencies, where each layer's solution creates the preconditions for the next:
omni-modal data introduces workload heterogeneity that demands service-level fault isolation;
once RL roles run as independent services, a dedicated asynchronous data bus is needed to prevent synchronous data exchange from propagating delays across roles;
and the field-based storage model of the asynchronous bus naturally accommodates heterogeneous modality fields, closing the loop back to the first challenge.

Guided by this causal progression, we make the following contributions:

\textbf{1) Role-Isolated Service Architecture} (\S\ref{sec:server-arch}).
We introduce a service deployment model in which each RL role runs as an independent, fault-isolated service built on Ray Serve~\cite{moritz2018ray}.
The design delivers three core operational benefits: fault isolation via two-tier recovery, independent scaling of individual roles, and role-level lifecycle management.
A self-developed Distributed Checkpoint Service (DCS) provides efficient weight synchronization across heterogeneous cluster topologies.

\textbf{2) Staleness-Unified Asynchronous Training} (\S\ref{sec:async-training}).
We integrate TransferQueue~\cite{han2025asyncflow}, an open-source distributed data bus, and adapt it for RL training.
Our adaptations include a staleness parameter that unifies on-policy and off-policy training under a single knob, and a streaming micro-batch pipeline that eliminates long-tail blocking by allowing downstream stages to consume data as it arrives.
The result is a system where the same codebase, the same entry point, and a single configuration parameter control whether training runs synchronously, near-on-policy, or fully asynchronously.

\textbf{3) Omni-Modal Agentic RL} (\S\ref{sec:omni-native}).
We design a unified multimodal data preprocessing pipeline that natively handles images, text, audio, and video, with modality-aware parallel strategies including ViT replication across tensor-parallel ranks and encoder-aware pipeline placement.
Relax further provides clean extension points for agentic RL workflows, including multi-turn rollout, custom reward services, and tool/sandbox integration (\S\ref{sec:agentic}).
We demonstrate verified end-to-end RL training convergence for Qwen3-Omni across image, text, audio, and video.

We validate these contributions in \S\ref{sec:experiments}.
On a 16$\times$H800 cluster, Relax's fully asynchronous off-policy mode achieves a 1.76$\times$ speedup over colocate training, and its on-policy mode achieves 1.12$\times$ over colocate.
We further demonstrate stable RL training convergence on Qwen3-Omni-30B across image, text, audio, and video modalities over 2{,}000 steps.

\section{Background and Related Work}
\label{sec:background}

\subsection{RL Algorithms for LLM Alignment}
\label{subsec:rl-algorithms}

Reinforcement learning from human feedback (RLHF) was first proposed to align language models with human preferences by training a reward model and optimizing against it with policy gradient methods~\cite{christiano2017rlhf}.
InstructGPT~\cite{ouyang2022instructgpt} demonstrated the practical viability of this approach at scale, establishing RLHF as a core post-training technique.
Proximal Policy Optimization (PPO)~\cite{schulman2017ppo} became the default algorithm due to its stable clipped surrogate objective, but its reliance on a learned reward model and a separate value network imposes significant memory and compute overhead.
Group Relative Policy Optimization (GRPO)~\cite{shao2024deepseekmath} addressed this by removing the value network entirely, instead estimating baselines from group-level statistics over sampled responses, thereby cutting memory consumption and simplifying the training pipeline.
More recently, DAPO~\cite{yu2025dapo} introduced decoupled clip ratios and dynamic sampling strategies that improve training stability at scale.

In parallel, \emph{RL with Verifiable Rewards} (RLVR) has emerged as a powerful paradigm in which reward signals come from programmatic verifiers, such as unit tests for code or symbolic checkers for mathematics, rather than learned reward models.
DeepSeek-R1~\cite{deepseek2025r1} demonstrated that RLVR can incentivize emergent reasoning capabilities, including self-reflection and step-by-step verification, without requiring human-labeled trajectories.
A central tension across these methods is the on-policy versus off-policy tradeoff, which we revisit from a systems perspective in \S\ref{sec:async-training}.

\subsection{Agentic RL}
\label{subsec:agentic-rl}

Beyond single-turn prompt--response RL, a growing body of work applies reinforcement learning to \emph{agentic} settings where the model interacts with external environments across multiple turns.
These settings include multi-turn reasoning in which the model iteratively refines its chain of thought~\cite{chen2025longcot}; tool use that involves invoking APIs, calculators, or code interpreters mid-generation; and search-augmented reasoning with retrieval calls interleaved with generation steps.
ProRL~\cite{zhang2026prorl} formalized the rollout-as-a-service abstraction for multi-turn agentic RL, decoupling sandbox environments from the training loop.

Agentic RL imposes qualitatively new requirements on training engines, including variable-length trajectories from unpredictable environment interactions, extended context windows from long chain-of-thought reasoning~\cite{chen2025longcot}, and highly variable environment response latencies (cf.\ the challenges outlined in \S\ref{subsec:challenges}).
Existing work has begun to address these requirements from complementary angles: ProRL decouples sandbox environments from the training loop via a rollout-as-a-service abstraction, while OpenRLHF~\cite{hu2024openrlhf} combines asynchronous RL with agentic multi-turn workflows.
Relax builds on these ideas through its service-oriented architecture (\S\ref{sec:server-arch}) and asynchronous data bus (\S\ref{sec:async-training}).

\subsection{Existing RL Training Systems}
\label{subsec:existing-systems}

The rapid growth of RL post-training has produced a diverse ecosystem of open-source training frameworks, each advancing the state of the art along different dimensions.
We briefly survey representative systems below; all frameworks are rapidly evolving, and the descriptions reflect publicly available information as of April 2026.

\paragraph{Dataflow and programming model.}
A central design question for RL training systems is how to manage data flow between heterogeneous roles (actor, critic, reward model, reference model).
veRL~\cite{sheng2024hybridflow} unifies SPMD computation with RPC-based data transfer in its HybridFlow programming model, enabling flexible role placement and 3D-HybridEngine resharding.
OpenRLHF~\cite{hu2024openrlhf} prioritizes ease of use and extensibility through a Ray-based scheduling architecture with DeepSpeed integration, lowering the barrier for researchers to experiment with diverse RL algorithms.
AReaL~\cite{fu2025areal} targets large-scale asynchronous training, introducing staleness-aware policy optimization to manage the freshness--throughput tradeoff.
NeMo-RL, whose rollout infrastructure is described in ProRL Agent~\cite{zhang2026prorl}, provides tight Megatron-native integration for service-oriented RL training.
Slime~\cite{slime2025} combines Megatron for 3D-parallel training with SGLang for inference serving, bridging the two through a centralized data buffer that manages prompt scheduling and rollout generation.

\paragraph{Active progress in omni-modal RL.}
Vision-language model (VLM) RL training is actively supported by multiple frameworks.
veRL supports vision-language models such as Qwen3-VL, with ongoing work toward broader multimodal coverage; ROLL~\cite{wang2025roll} provides user-friendly abstractions for VLM-scale RL.
However, audio and video modality RL training remains at an early stage across the community.

\paragraph{Training mode flexibility.}
The ability to flexibly switch between on-policy, near-on-policy, and off-policy training modes requires careful system-level abstraction.
AReaL implements a mature asynchronous training architecture with staleness-enhanced PPO, demonstrating that controlled off-policy training can achieve significant speedups without convergence degradation.
OpenRLHF combines asynchronous RL with agentic workflows, enabling multi-turn rollouts to proceed concurrently with training.

Relax integrates these three dimensions into a unified architecture: a service-oriented deployment with independent fault domains and elastic scaling (\S\ref{sec:server-arch}), an asynchronous data bus whose single staleness parameter governs the full on-policy-to-off-policy spectrum (\S\ref{sec:async-training}), and an omni-native pipeline with modality-aware parallelism and agentic RL extensibility (\S\ref{sec:omni-native}).
The system is built on Ray~\cite{moritz2018ray} for orchestration, Megatron~\cite{shoeybi2019megatron} for 3D-parallel training, and SGLang~\cite{zheng2023sglang} for inference serving.

\section{System Architecture}
\label{sec:overview}

\begin{figure}[htbp]
  \centering
  \includegraphics[width=\linewidth]{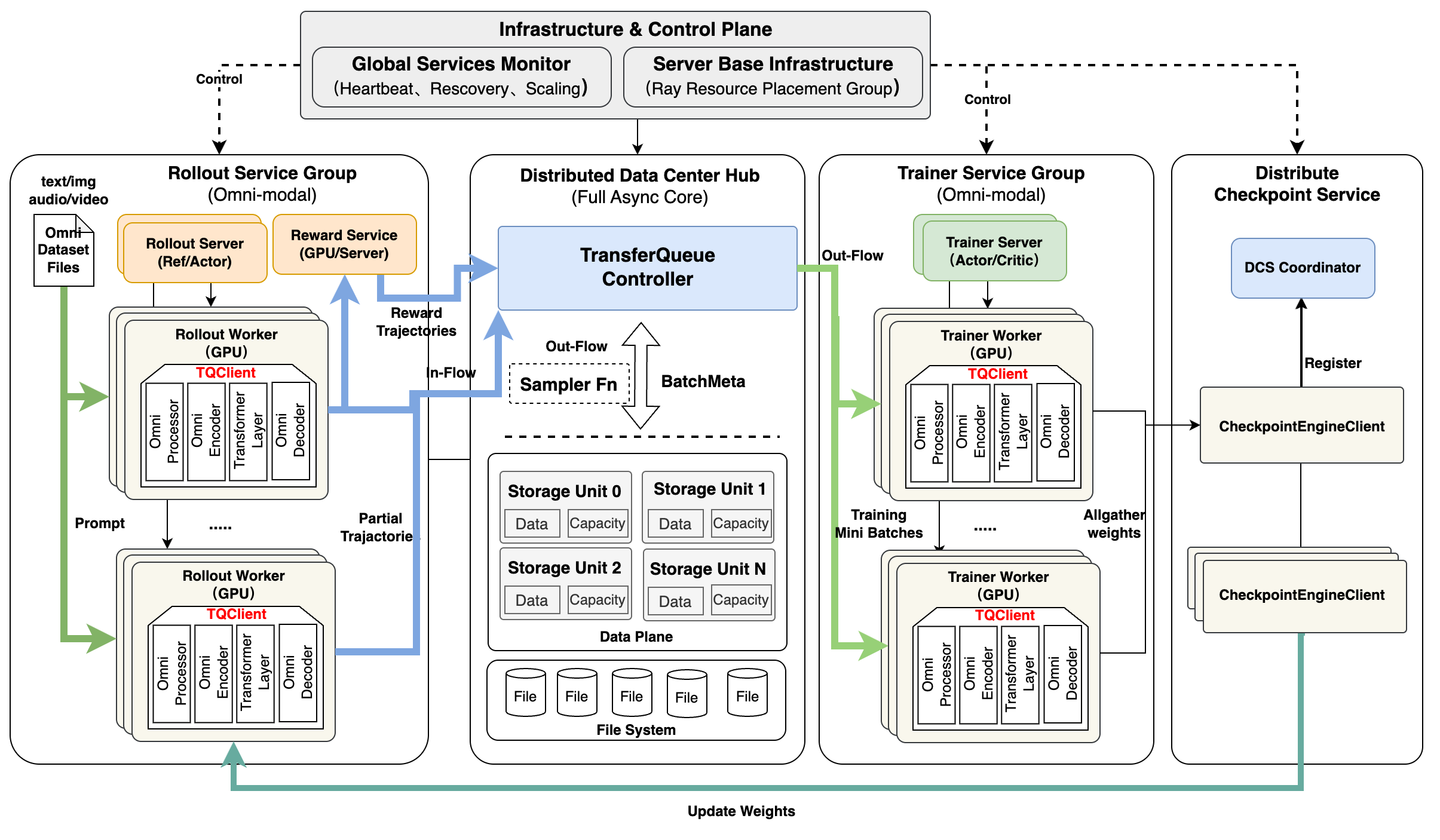}
  \caption{Relax system architecture. The Controller (control plane) orchestrates RL roles that execute on independent backends (computation plane) and exchange data exclusively through TransferQueue (data plane).}
  \label{fig:arch}
\end{figure}

A key design goal of Relax is to let each functional concern evolve, scale, and fail independently.
This section describes the resulting decomposition (\S\ref{subsec:architecture}) and the two execution modes it enables.

\subsection{Overview}
\label{subsec:architecture}

As illustrated in Figure~\ref{fig:arch}, Relax decomposes the RL training system into three orthogonal planes:

\begin{enumerate}
  \item \textbf{Control plane.}
    The Controller orchestrates the training loop by issuing high-level directives (e.g., ``generate rollouts,'' ``compute advantages,'' ``run a gradient step'') without embedding any computation logic itself.

  \item \textbf{Computation plane.}
    Each RL role executes its workload inside an independent backend.
    Training roles use Megatron-LM for distributed model parallelism; inference roles use SGLang for high-throughput generation; reward models may use either backend depending on the task.

  \item \textbf{Data plane.}
    A \emph{TransferQueue} (TQ) data bus mediates all inter-role data movement.
    All components exchange data exclusively through TQ; no role holds a direct reference to another.
    A separate checkpoint service (DCS, detailed in \S\ref{sec:server-arch}) handles weight synchronization between training and inference replicas.
\end{enumerate}

This three-way separation yields a key engineering benefit: any plane can be upgraded independently.
For example, replacing the training backend (e.g., migrating from Megatron-LM to FSDP~\cite{zhao2023fsdp}) requires no changes to the data bus or orchestration logic, and scaling the inference cluster does not affect the training loop.

Relax supports two primary execution modes, unified under a single codebase.
In \textbf{colocate mode}, training and inference share the same GPU cluster and alternate execution, maximizing GPU memory utilization at the cost of serializing the two workloads.
In \textbf{fully asynchronous mode}, training and inference run on independent GPU clusters and exchange data exclusively through the TransferQueue.
Because neither cluster waits for the other, training steps and rollout generation overlap almost completely, yielding substantially higher throughput (see \S\ref{sec:experiments}).
\subsection{Service-Oriented Architecture}
\label{sec:server-arch}

The ideal operational experience for large-scale RL training resembles that of serverless computing: each functional role in the training pipeline should be independently deployable, elastically scalable, and individually recoverable without disturbing unrelated roles.
However, GPU-resident RL workloads are fundamentally at odds with the serverless execution model: they are long-running, stateful (model weights, optimizer states, distributed communication groups), and require persistent GPU memory allocations that preclude the ephemeral, event-driven invocation pattern of serverless functions.

Inspired by the operational simplicity of serverless platforms, Relax adopts a \emph{service-oriented architecture} that accommodates the persistent, stateful nature of GPU workloads while delivering serverless-like operational benefits.
Each RL role is deployed as an independent Ray Serve Deployment~\cite{moritz2018ray} with its own failure domain, resource quota, and health monitoring.
The Controller registers role services at startup based on the chosen algorithm (e.g., GRPO~\cite{shao2024deepseekmath}, DAPO~\cite{yu2025dapo}) and orchestrates them without embedding their execution logic.
Compared with monolithic training loops adopted by prior frameworks~\cite{sheng2024hybridflow,hu2024openrlhf,fu2025areal}, this design provides three concrete advantages:
\textbf{(i)}~faults are contained at the service boundary, so that a crash in one role (e.g., an out-of-memory error in the reward model) does not propagate to others;
\textbf{(ii)}~roles with heterogeneous resource demands can be scaled independently, for example adding rollout replicas without resizing the critic cluster;
and \textbf{(iii)}~each role's lifecycle, from initialization through checkpointing to restart, is managed at the service level rather than entangled with a global training loop.
Two deployment configurations are supported: in the \emph{colocate mode} (default), the Actor, Critic, and Rollout roles share GPU resources via time-multiplexed scheduling; in the \emph{fully asynchronous mode}, the system deploys six independent roles on dedicated resources to maximize training and inference overlap (\S\ref{sec:async-training}).

\begin{figure*}[htbp]
  \centering
  \begin{subfigure}[b]{0.48\textwidth}
    \centering
    \includegraphics[width=\textwidth]{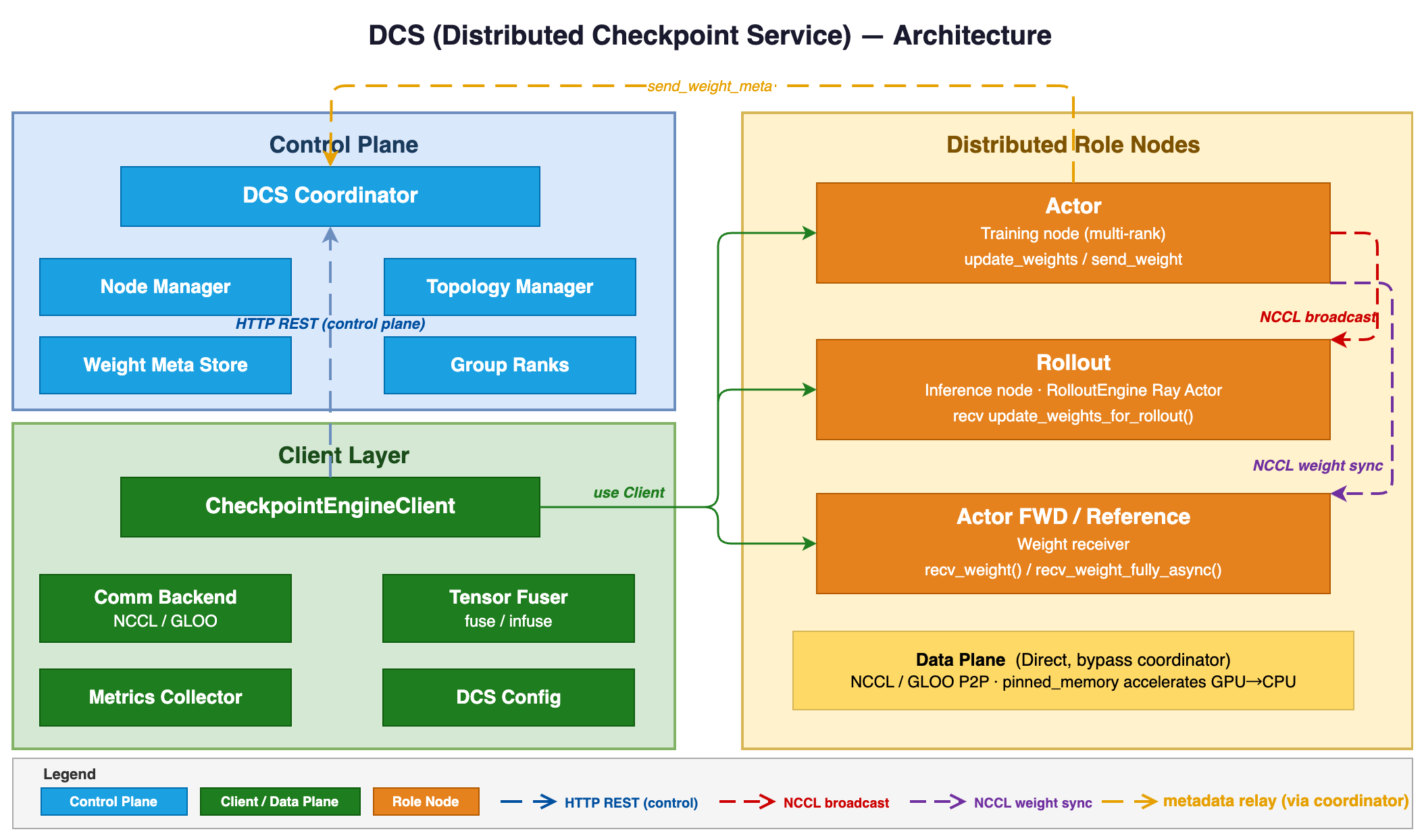}
    \caption{Distributed Checkpoint Service (DCS) architecture.}
    \label{fig:dcs-arch}
  \end{subfigure}
  \hfill
  \begin{subfigure}[b]{0.48\textwidth}
    \centering
    \includegraphics[width=\textwidth]{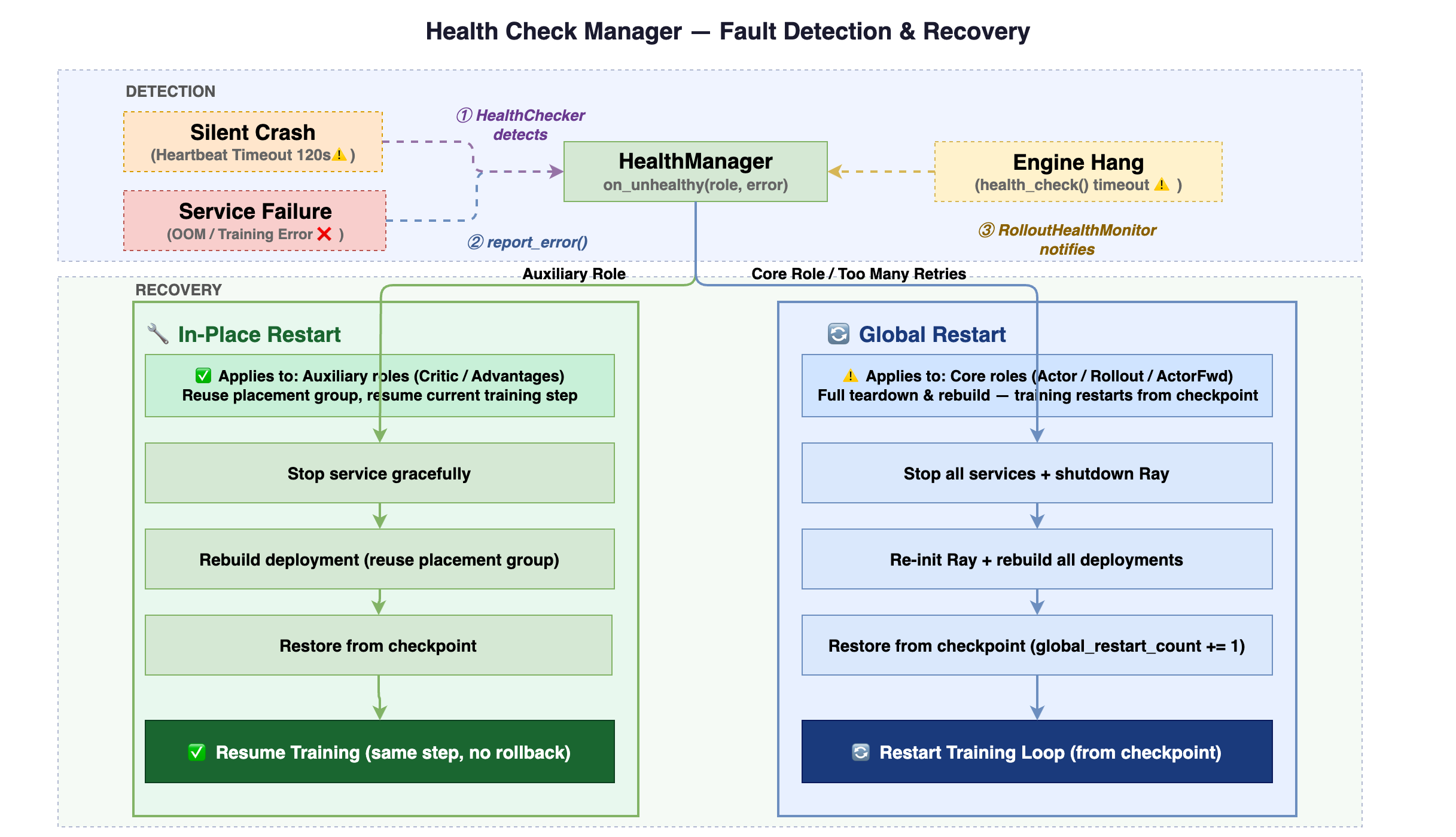}
    \caption{Fault detection and recovery flow.}
    \label{fig:fault-recovery}
  \end{subfigure}
  \caption{Service-oriented infrastructure components. (a)~DCS coordinates topology discovery and weight metadata; clients abstract transport details; role nodes perform low-latency weight synchronization. (b)~The health subsystem classifies role failures by severity and triggers either in-place or global restart accordingly.}
  \label{fig:service-infra}
\end{figure*}

\paragraph{Distributed Checkpoint Service (DCS).}
\label{subsec:dcs}
A latency-critical operation in RL training is the synchronization of updated model weights from the training engine to all inference engines after each policy update.
Modern RL frameworks typically perform this synchronization via in-memory resharding within training or inference processes.
Relax extracts weight synchronization into a dedicated, independently deployed service---the \emph{Distributed Checkpoint Service} (DCS), which distributes updated weights to inference engines with low latency as a standalone Ray Serve deployment.

The DCS architecture comprises two core components.
The \textbf{Coordinator}, deployed as a Ray Serve application, serves as the control plane: it performs topology discovery, coordinates synchronization barriers across roles, and exposes Prometheus metrics for observability.
The \textbf{CheckpointEngineClient} provides a unified interface for saving, loading, and registering checkpoints, abstracting away transport details from role implementations; internally, it embeds a TopologyManager for discovering heterogeneous TP/PP rank mappings and a DCS Protocol layer for efficient tensor serialization.
DCS supports two transport backends selected based on deployment topology:
the \textbf{NCCL-based backend} for intra-cluster GPU-to-GPU transfer, where training TP shards are gathered via NCCL AllGather and broadcast to all inference engine ranks, achieving the lowest synchronization latency by keeping data entirely on GPU memory;
and the \textbf{TCP-based backend} for cross-cluster scenarios where NCCL connectivity is unavailable, offloading tensors from GPU to CPU memory for TCP transfer---essential for elastic scaling where additional Rollout replicas may be provisioned in a separate cluster.

\paragraph{Health monitoring and fault tolerance.}
\label{subsec:health-monitor}
Long-running distributed RL training is inherently susceptible to partial failures: GPU memory fragmentation, NCCL timeout, inference engine hangs, and transient network partitions are routine rather than exceptional at scale.
A monolithic training loop typically responds to any such failure with a full-job restart, wasting hours of computation.
Relax instead provides fine-grained, service-level fault tolerance through a dedicated health monitoring subsystem comprising four components:
a per-service health state record that tracks status, heartbeat timestamps, failure counts, and recovery history;
a global health store, implemented as a Ray remote Actor, that aggregates state from all services;
a monitoring daemon that periodically probes each service via lightweight heartbeats;
and a Controller-facing manager that consumes health state transitions and triggers recovery actions.
Not all role failures are equal.
Stateless or recomputable roles (Critic, Advantages) can be restarted in place: the service process is killed and relaunched on the same resources, with no impact on other roles.
In contrast, roles that carry authoritative model weights or participate in distributed communication groups (Actor, Rollout, ActorFwd) require a \emph{global restart}, tearing down Ray Serve deployments, TransferQueue partitions, DCS channels, and Ray Actor Groups, after which the Controller re-initializes from the last valid checkpoint.
\subsection{Asynchronous Training}
\label{sec:async-training}

Service decoupling (\S\ref{sec:server-arch}) does not imply execution decoupling: if services exchange intermediate results through synchronous RPCs or shared-memory buffers, a delay in any single stage still propagates to every downstream consumer.
To fully realize the promise of a service-oriented architecture, the data exchange mechanism itself must be asynchronous.

This motivates integrating \textbf{TransferQueue} (TQ)~\cite{han2025asyncflow}\footnote{TransferQueue is an open-source project developed by Ascend: \url{https://github.com/Ascend/TransferQueue}. Relax integrates TQ as an external dependency; our contribution is the RL-specific adaptation layer described below.}, an independent, asynchronous data bus through which all Relax services exchange training data via a unified client API.
A key property that makes TQ well-suited for RL is \emph{field-based storage}: different fields of the same sample (e.g., generated responses, log probabilities, rewards) can be independently written and read, directly matching the multi-stage computation pattern where each RL stage produces a disjoint set of fields at different times.
Readers are referred to~\cite{han2025asyncflow} for a comprehensive treatment of TQ's internal design.

\begin{figure}[htbp]
  \centering
  \includegraphics[width=\columnwidth]{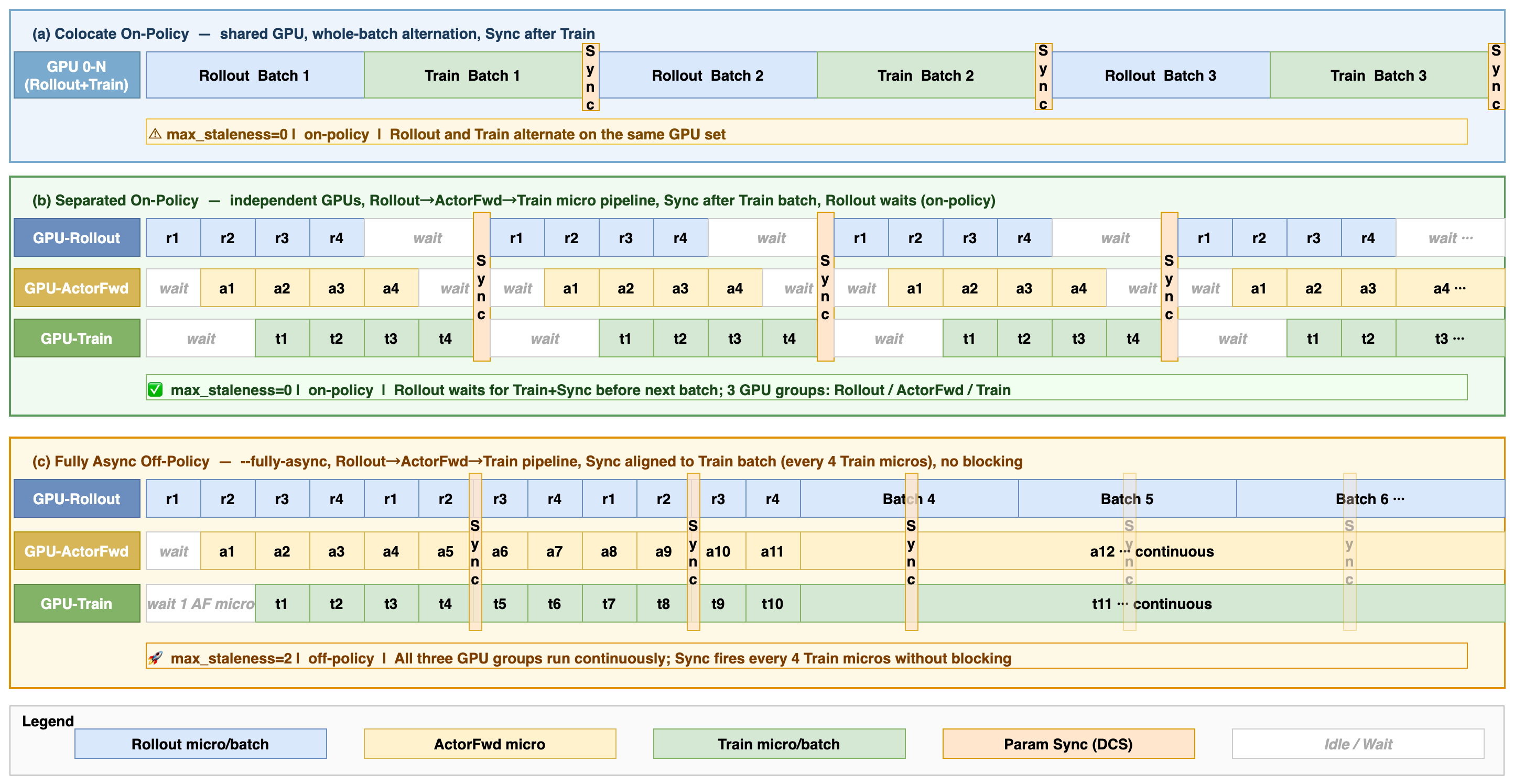}
  \caption{Timeline comparison of three training modes. (a)~Colocate: rollout and training alternate on shared GPUs. (b)~Separated on-policy: independent GPU groups with inter-batch synchronization. (c)~Fully async off-policy: all groups run continuously with no blocking.}
  \label{fig:async-compare}
\end{figure}

\paragraph{Unified training modes via staleness control.}
\label{subsec:staleness}
A fundamental design choice in RL training is the degree of \emph{policy staleness} tolerated between data generation and policy update.
On-policy algorithms (e.g., PPO~\cite{schulman2017ppo}, GRPO~\cite{shao2024deepseekmath}) require that training data be generated by the current policy, while off-policy methods allow reuse of data from older policy versions, trading policy freshness for higher throughput~\cite{fu2025areal}.
Rather than requiring separate code paths or configuration templates for each regime, Relax unifies the entire spectrum through a single integer parameter: \texttt{max\_staleness}.
Let $v_t$ denote the current weight version on the training side, and let $v_r$ denote the weight version used by the rollout worker when it generated a given batch of data.
We define \emph{staleness} as the version gap $s = v_t - v_r$.
The rollout worker continues generating new data only when $s < \texttt{max\_staleness}$; otherwise, it pauses until the training side consumes older partitions and the gap decreases, thereby self-regulating the producer--consumer balance without centralized coordination.
Crucially, the staleness check executes on the \emph{rollout side} (the data producer), not on the Controller or the training side, ensuring that the producer self-regulates without flooding the data bus with increasingly off-policy samples.
The key operational benefit is that the \emph{same codebase and entry script} supports all three modes---users only adjust the staleness threshold, eliminating the need for separate synchronous and asynchronous training configurations.

\paragraph{Streaming data flow.}
\label{subsec:streaming-dataflow}
Staleness control determines \emph{how asynchronous} training is allowed to be; the streaming data flow layer determines \emph{how} data actually moves between producers and consumers to realize that asynchrony.
Traditional RL systems adopt a \emph{global-batch-synchronous} scheduling model: the rollout engine generates an entire global batch (e.g., 256 samples), and only after every sample is complete does the system dispatch the batch to downstream stages.
This design introduces a \emph{long-tail bottleneck}: a single sample with an unusually long response (e.g., a 20k-token chain-of-thought output~\cite{chen2025longcot}) blocks the entire step, leaving all downstream workers idle.
Relax replaces global-batch synchronization with \emph{streaming micro-batch scheduling}: the rollout engine partitions the global batch into micro-batches (e.g., 32 samples each), and as soon as a micro-batch completes generation, it is immediately written to TQ as a ready partition for downstream consumption.
On the training side, a streaming data loader continuously monitors partition-ready events and yields micro-batches to the training engine as they become available, supporting both PP rank-0 broadcast (colocate mode) and independent per-stage pulls (fully async mode).
Together, micro-batch scheduling and the streaming data loader form a complete producer--consumer pipeline that directly explains the \textbf{0.1\% trainer idle ratio} observed in fully asynchronous mode (see \S\ref{sec:experiments}).

\paragraph{Multimodal field-level decoupling.}
The streaming data flow extends naturally to multimodal RL training through TQ's field-based storage.
Relax organizes multimodal sample fields into four categories---text, image, audio, and video---each with distinct sizes and preprocessing latencies.
Because TQ tracks readiness at the field level, different modality fields can be written to the data bus at different times without blocking each other.
For instance, image fields may be ready within milliseconds while video decoding for the same sample takes seconds; downstream services that depend only on text and image fields can begin processing immediately without waiting for video fields.
This field-level decoupling connects seamlessly with the omni-modal pipeline described in \S\ref{sec:omni-native}, where each modality has its own preprocessing and encoding path.
\section{Omni-Modal Agentic RL}
\label{sec:omni-native}

The system architecture described in \S\ref{sec:overview} provides a modular, asynchronous infrastructure backbone.
This section describes how Relax builds upon that backbone to address two frontier demands: native support for omni-modal training (\S\ref{subsec:omni-pipeline}) and extensibility for agentic RL workflows (\S\ref{sec:agentic}).

Existing RL training frameworks are predominantly designed around text-only workloads.
When extended to multimodal inputs, they typically bolt on ad hoc preprocessing wrappers that convert images or audio into token-like representations, apply text-centric parallel strategies that ignore the structural differences of vision encoders, and rely on inference backends that do not natively handle heterogeneous media batches.
These adaptations are fragile, hard to generalize across modalities, and leave significant performance on the table.
Relax instead adopts an \emph{omni-native} design philosophy: multimodal support is a first-class design constraint that permeates every layer of the system---from data ingestion, through parallel execution, to model onboarding.

\subsection{Omni-Modal Pipeline Design}
\label{subsec:omni-pipeline}

Relax's omni-modal pipeline spans three stages: a unified data pipeline that ingests heterogeneous media, omni-modal-aware parallel strategies, and integration with NVIDIA's Megatron Bridge for checkpoint conversion.
We describe each in turn.

\paragraph{Unified data pipeline.}
\label{subsec:mm-data-pipeline}
A central challenge of omni-modal RL is that raw training samples contain heterogeneous media---images, video clips, and audio segments---interleaved with text within multi-turn conversations.
Relax addresses this with a unified data pipeline that transforms raw JSON records with media references into model-ready batch tensors.
Each sample's media references are resolved via modality-specific loaders that produce typed modality objects with standardized metadata (resolution, duration, sample rate).
A dedicated preprocessing stage arranges all modality objects in conversation turn order and assigns a binary mask to each element, distinguishing user-provided inputs from assistant-generated outputs---critical for RL training, where the policy loss must be computed only over assistant tokens while conditioning on all preceding modalities.
Modality-specific processors then convert parsed objects into tensor representations (pixel values, video frame sequences, or audio features), and the chat template encoder produces token sequences with interleaved multimodal placeholders, together with labels and attention masks.
The collator handles heterogeneous padding across samples with different numbers and types of media, computes cumulative sequence lengths for flash attention, and performs sequence-parallel slicing when needed.
A key design choice is \emph{dynamic transform injection}: the encoding function is assembled via partial application with model-specific processor and template arguments, enabling the same pipeline to serve different model families (e.g., Qwen-VL, Qwen-Omni) by swapping only the processor and template components.

\paragraph{Multimodal parallel strategies.}
\label{subsec:mm-parallel}
Vision-language models introduce structural heterogeneity that standard LLM parallel strategies do not account for.
Relax provides two co-designed parallel strategies.
\emph{ViT Tensor Parallelism} replicates the vision encoder on all TP ranks rather than partitioning its parameters, treating the ViT as data-parallel within the TP group; an AllReduce merges visual features before they enter the language backbone, with negligible overhead since ViT parameters are typically 1--5\% of the total model size.
\emph{Encoder-Aware Pipeline Parallelism} places all modality-specific encoders (e.g., ViT for vision, Whisper for audio) on the first pipeline stage (PP0), so that bulky raw inputs are consumed locally without crossing stage boundaries, minimizing inter-stage communication and avoiding load imbalance from encoder layers with substantially different FLOPs profiles.

\paragraph{Model coverage and Megatron Bridge.}
\label{subsec:model-coverage}
To support a broad range of omni-modal models atop Megatron-LM~\cite{shoeybi2019megatron}, Relax builds upon NVIDIA's open-source \emph{Megatron Bridge}\footnote{\url{https://github.com/NVIDIA-NeMo/Megatron-Bridge}} to provide automatic bidirectional weight conversion between HuggingFace and Megatron checkpoint formats.
On top of the community bridge, Relax extends support for omni-modal-specific components---including MoE expert sharding and multi-modal encoder weight mapping (e.g., ViT, Whisper)---enabling practitioners to start from any publicly released HuggingFace checkpoint and seamlessly train with Megatron's optimized 3D parallelism.
Upon completion, weights are converted back to the HuggingFace format for downstream deployment and evaluation via SGLang~\cite{zheng2023sglang} or other inference engines.
\subsection{Extensibility for Agentic RL}
\label{sec:agentic}

A key design goal is to keep infrastructure concerns and algorithm concerns cleanly separated: the training infrastructure (service orchestration, data transport, weight synchronization) is managed by Relax, while algorithm researchers define rollout logic, reward functions, and tool integrations in an independent recipe workspace.
Relax abstracts these extension points behind unified interfaces, so that researchers can develop and iterate on agentic RL recipes without modifying the distributed training internals.

\paragraph{Custom rollout and reward.}
Relax supports multi-turn agentic workflows where each reasoning turn may receive new visual inputs (e.g., image--text interleaved dialogues).
The service-oriented architecture makes multi-turn state management straightforward: rollout services maintain per-session conversation state across turns, while TQ tracks field-level readiness for each turn independently.
Reward computation is treated as a pluggable service with two built-in backends and an open extension point.
\emph{Rule-based rewards} apply deterministic scoring functions (e.g., format checking, answer verification) directly within the rollout loop.
\emph{Generative reward models} (GenRM) deploy an LLM-as-Judge~\cite{liu2025deepseekgrm} as an independent Ray Serve service, preserving service isolation: the GenRM scales, fails, and recovers independently of the policy training loop.
For scenarios requiring domain-specific evaluation, researchers implement a custom reward interface and register it as a service endpoint; Relax handles scheduling, batching, and fault tolerance automatically.

\paragraph{Tool and sandbox integration.}
Relax accommodates external tool calling and search-augmented reasoning within the rollout loop.
Tool invocations are treated as asynchronous service calls, fitting naturally into the service-oriented data flow without requiring modifications to the training pipeline.
For agentic tasks that require isolated execution environments (e.g., code execution, web browsing), Relax provides a sandbox integration interface: researchers define the sandbox lifecycle (setup, execute, teardown) and Relax manages sandbox instances as ephemeral services, ensuring resource isolation and automatic cleanup.
\section{Experiments}
\label{sec:experiments}

We evaluate Relax along four dimensions that correspond to its core design claims: \emph{omni-modal convergence} across image, audio, video, and agentic modalities (\S\ref{subsec:exp-convergence}), \emph{end-to-end performance} against veRL under an identical technology stack (\S\ref{subsec:exp-perf}), \emph{training-mode tradeoffs} among colocate, on-policy, and off-policy execution (\S\ref{subsec:exp-modes}), and \emph{training stability} on MoE models where rollout-training routing mismatch can silently degrade optimization (\S\ref{subsec:exp-stability}).

\subsection{Experimental Setup}
\label{subsec:exp-setup}

Our experiments span four model families covering dense, MoE, and omni-modal architectures.
For omni-modal convergence, we train Qwen3-Omni-30B on Echo Ink (image+text+audio) and NextQA (video), with supplementary text-only (DAPO-MATH-17k~\cite{yu2025dapo}, Qwen3-30B-A3B) and agentic (Deepeyes, Qwen3-VL-MoE-30B) validations in Appendix.
For end-to-end performance, we benchmark Qwen3-4B on DAPO-MATH-17k against veRL~\cite{sheng2024hybridflow}, which is configured with the same Megatron-LM~\cite{shoeybi2019megatron} + SGLang~\cite{zheng2023sglang} stack, in off-policy mode.
For training-mode comparison, we evaluate colocate, on-policy, and off-policy modes on both the 4B text-only and 30B omni-modal configurations to study how the throughput--freshness tradeoff scales with model size.
For training stability, we ablate R3 on Qwen3-30B-A3B (MoE, 30B total / 3B active) in a 2$\times$2 matrix (Relax/veRL $\times$ with/without R3).

\FloatBarrier
\subsection{Omni-Modal Convergence}
\label{subsec:exp-convergence}

\begin{figure*}[htbp]
  \centering
  \begin{subfigure}[b]{0.48\textwidth}
    \centering
    \includegraphics[width=\textwidth]{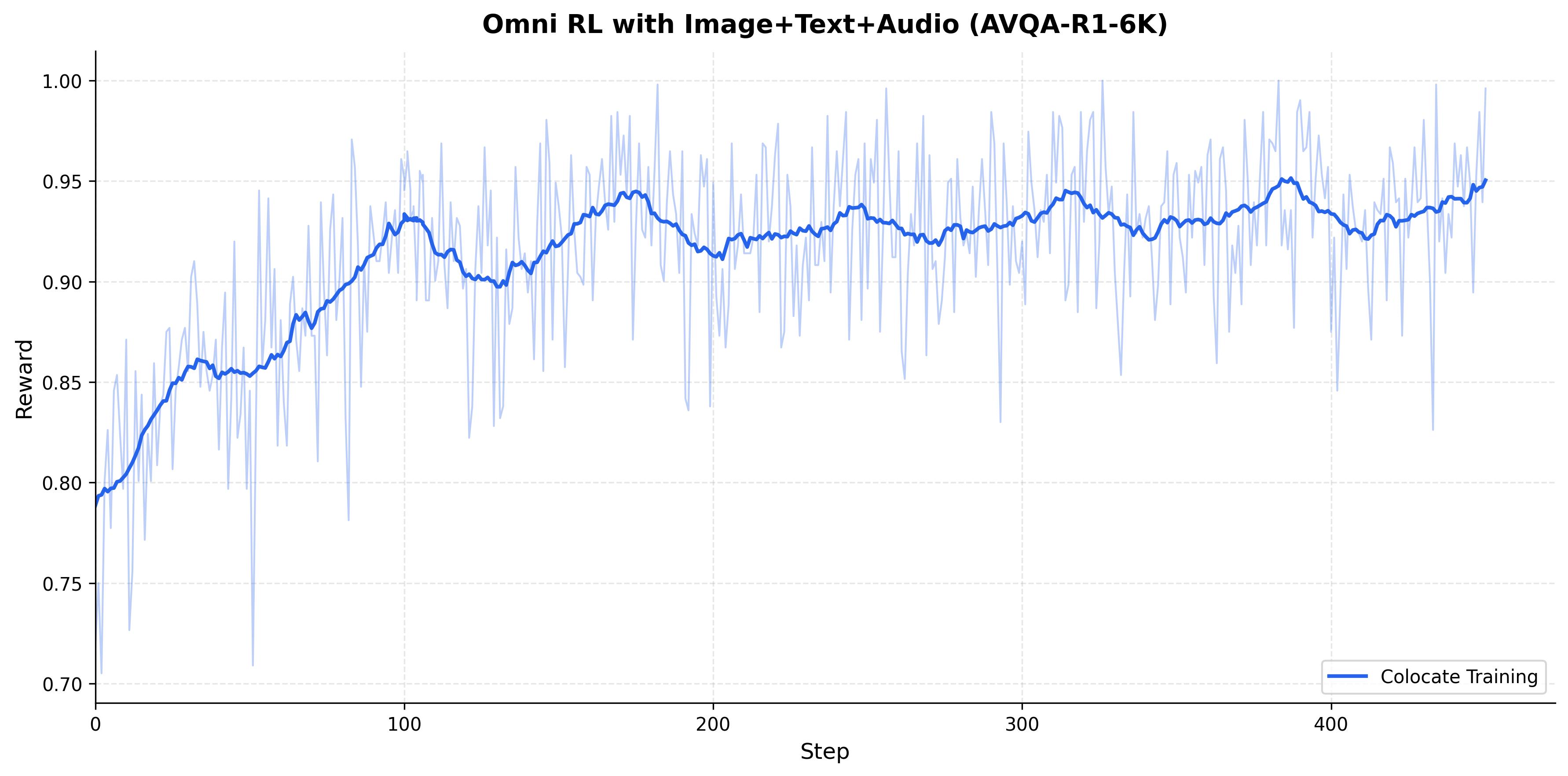}
    \caption{}
    \label{fig:conv-echoink}
  \end{subfigure}
  \hfill
  \begin{subfigure}[b]{0.48\textwidth}
    \centering
    \includegraphics[width=\textwidth]{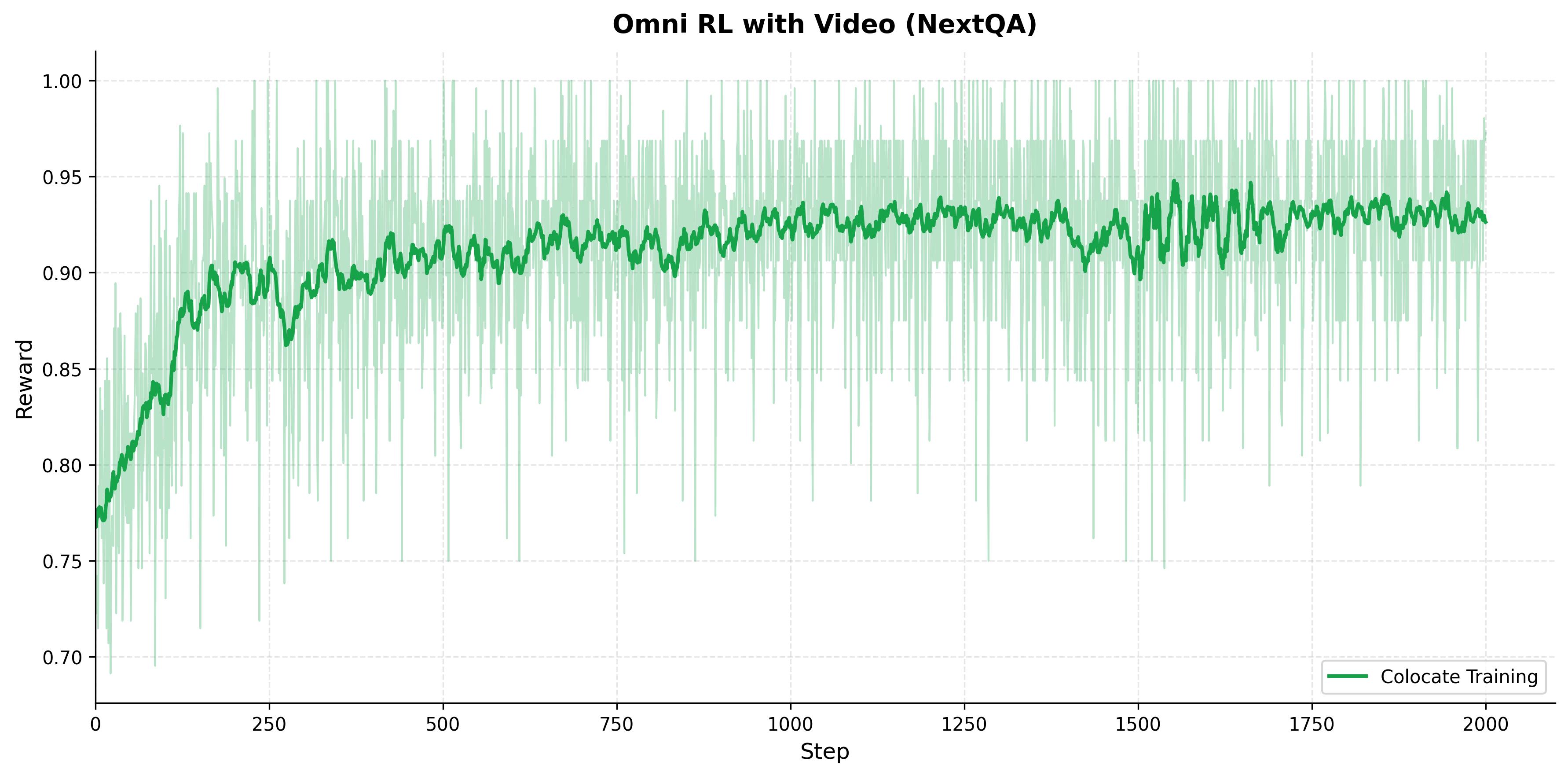}
    \caption{}
    \label{fig:conv-nextqa}
  \end{subfigure}
  \caption{Omni-modal reward convergence (Qwen3-Omni-30B). (a)~Image+text+audio training on Echo Ink. (b)~Video training on NextQA.}
  \label{fig:convergence-omni}
\end{figure*}

To validate that Relax produces correct training outcomes across diverse modality combinations, we replicate the EchoInk-R1 audio-visual reasoning setup~\cite{xing2025echoink} at a larger model scale.
Where the original work trains Qwen2.5-Omni-7B on the AVQA-R1-6K dataset (4{,}490 audio+image prompts) with GRPO, we substitute Qwen3-Omni-30B and train on 16$\times$H20 GPUs.
We then extend the evaluation to video understanding by training on a 0--30\,s subset of NextQA drawn from LLaVA-Video-178K.

On the Echo Ink audio+image task, Qwen3-Omni-30B converges from an initial reward of 0.72 to a plateau of ${\sim}0.93$ within 450 training steps (Figure~\ref{fig:conv-echoink}), validating that Relax's omni-native pipeline handles synchronized multi-modal inputs without framework-level modifications.
The NextQA video experiment further demonstrates long-run stability: over 2{,}000 continuous training steps the reward improves monotonically from ${\sim}$0.75 to ${\sim}$0.93 with no collapse or degradation (Figure~\ref{fig:conv-nextqa}), and variance remains consistently low (std $\approx$ 0.04--0.06) throughout---a critical property for production-scale omni-modal RL deployments.

While the primary focus of this section is omni-modal convergence, Relax is not limited to perception-only tasks.
On multi-turn agentic RL (Deepeyes, Qwen3-VL-MoE-30B), Relax matches veRL's convergence to the reward upper bound, confirming correct trajectory collection and advantage estimation under variable-length tool-calling episodes (Appendix~\ref{app:agentic-rl}).
On text-only DAPO-MATH, Qwen3-30B-A3B produces reward curves that align with veRL, with AIME-24 evaluation scores rising monotonically (Appendix~\ref{app:dapo-math}).
Together, these results demonstrate that Relax supports stable RL training across the full modality spectrum---from omni-modal perception to agentic multi-turn reasoning and pure text chain-of-thought.

\FloatBarrier
\subsection{End-to-End Performance}
\label{subsec:exp-perf}

We compare Relax against veRL on Qwen3-4B trained with DAPO~\cite{yu2025dapo} on DAPO-MATH-17k using 16$\times$H800 GPUs (maximum response length 20{,}480 tokens).
Both systems deploy training and inference on separate GPU nodes and operate in off-policy mode (staleness$=$1).
The key difference lies in pipelining granularity: Relax streams completed micro-batches to downstream stages immediately via TransferQueue, achieving micro-batch-level asynchrony, whereas veRL exchanges data at the global-batch level.

\begin{figure*}[htbp]
  \centering
  \begin{subfigure}[b]{0.48\textwidth}
    \centering
    \includegraphics[width=\textwidth]{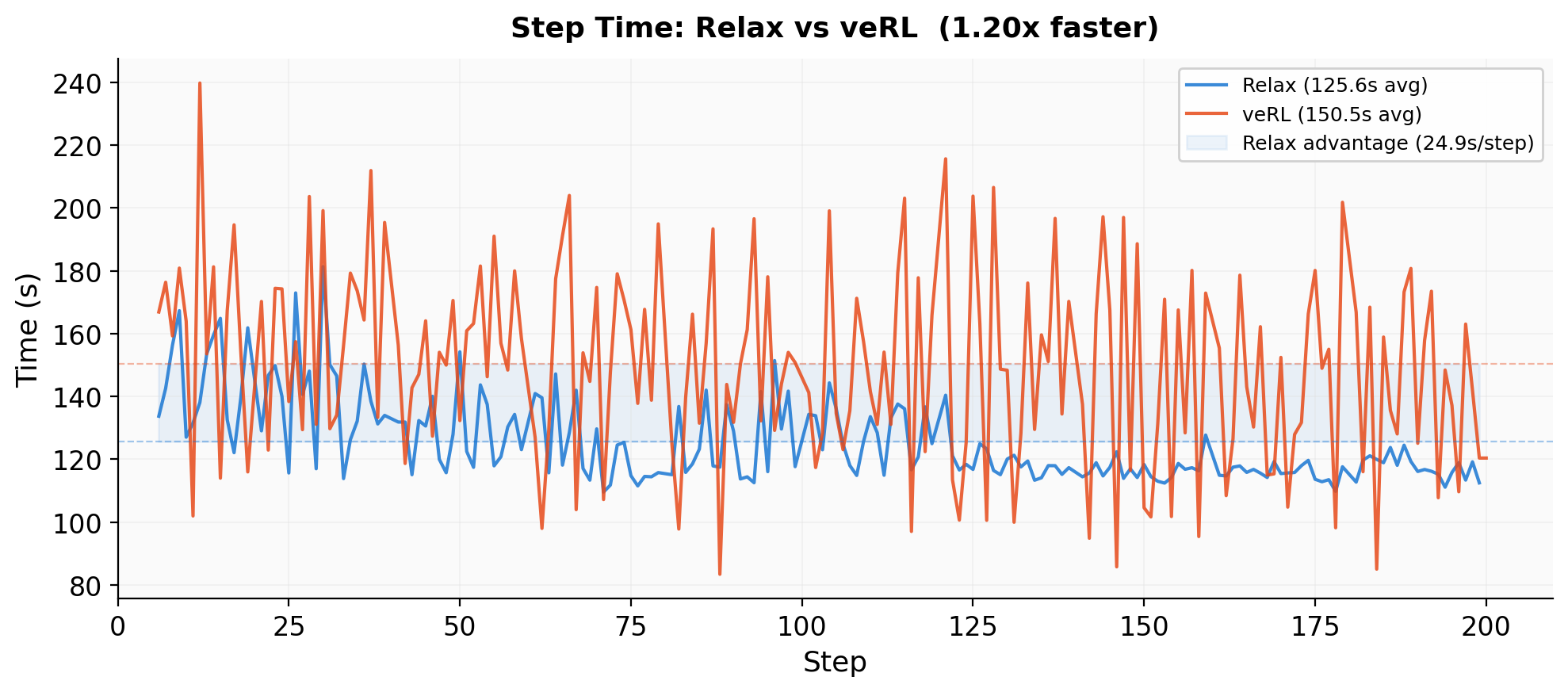}
    \caption{}
    \label{fig:perf-step-time}
  \end{subfigure}
  \hfill
  \begin{subfigure}[b]{0.48\textwidth}
    \centering
    \includegraphics[width=\textwidth]{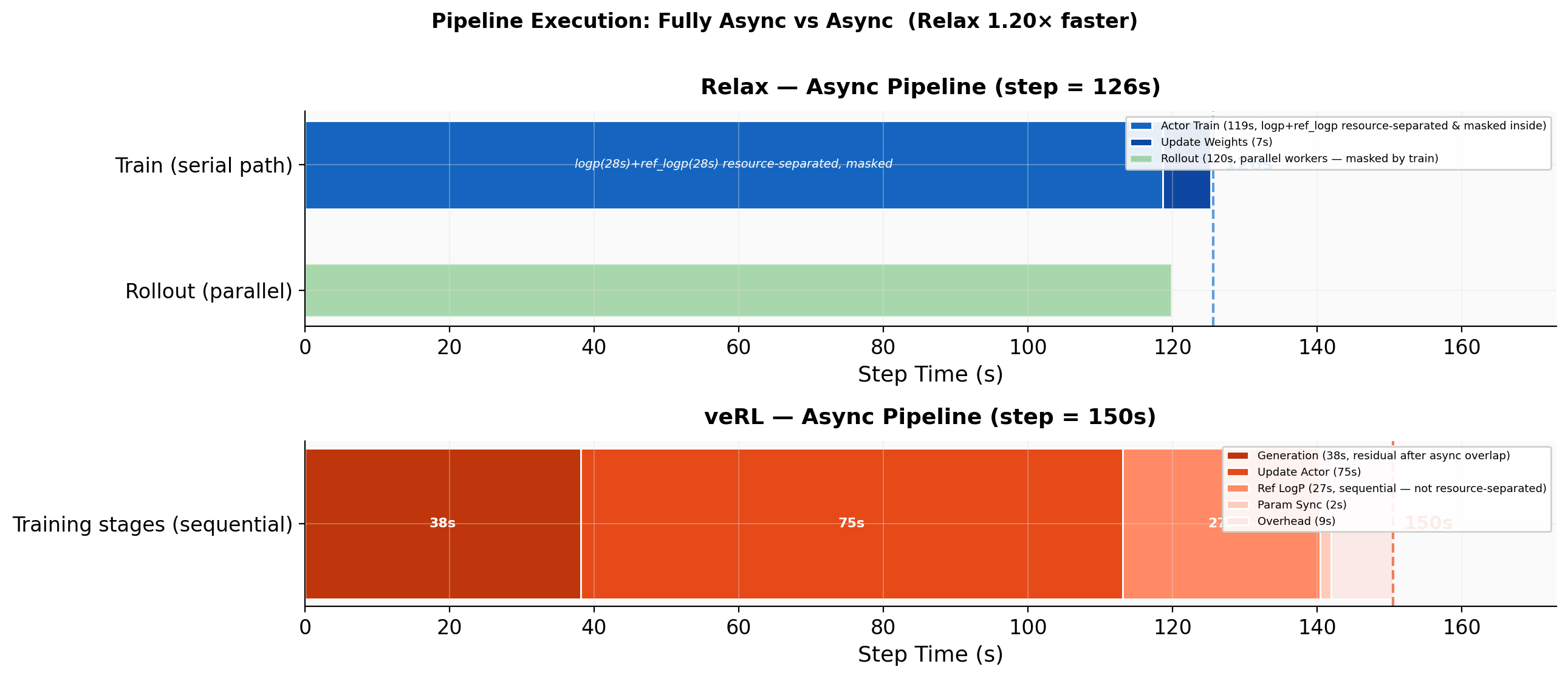}
    \caption{}
    \label{fig:perf-waterfall}
  \end{subfigure}
  \caption{End-to-end performance: Relax vs.\ veRL on Qwen3-4B. (a)~Per-step time comparison. (b)~Gantt-chart view of pipeline overlap.}
  \label{fig:relax-vs-verl-perf}
\end{figure*}

Relax achieves a \textbf{1.20$\times$ end-to-end speedup} (125.6\,s vs.\ 150.5\,s per step), translating to 28.7 vs.\ 23.9 steps/hour (Table~\ref{tab:relax-vs-verl}).
The per-step time chart (Figure~\ref{fig:perf-step-time}) confirms that this advantage is stable: Relax maintains a consistent ${\sim}$25\,s/step lead across all 195 effective training steps, with lower variance thanks to the micro-batch streaming mechanism that buffers long-tail generation latency.

\begin{table}[htbp]
\centering
\caption{End-to-end comparison: Relax vs.\ veRL on Qwen3-4B.}
\label{tab:relax-vs-verl}
\footnotesize
\begin{tabular}{@{}lll@{}}
\toprule
\textbf{Metric} & \textbf{Relax} & \textbf{veRL} \\
\midrule
Step Time              & \textbf{125.6\,s} & 150.5\,s \\
Steps/Hour             & \textbf{28.7}     & 23.9 \\
Rollout Wall-Clock Cost & 0\,s (fully masked) & 38.2\,s (on critical path) \\
Ref LogP Extra Cost    & 0\,s (resource separation) & 27.3\,s (sequential stage) \\
\bottomrule
\end{tabular}
\end{table}

The Gantt-chart view (Figure~\ref{fig:perf-waterfall}) reveals the structural source of this gap.
In Relax, rollout (119.8\,s) runs on separate inference nodes fully in parallel with training, and log-probability and reference log-probability computation (28.0\,s + 28.2\,s) are deployed on dedicated GPU resources via actor-train resource separation---a combined 56\,s of forward inference that contributes zero additional step time.
Because streaming micro-batch scheduling dispatches each completed micro-batch to downstream stages immediately, the training engine's data wait is only 0.11\,s per step, enabling virtually uninterrupted training even under high generation-latency variance.
In veRL, although rollout also runs on separate nodes, the generation residual (38.2\,s) and reference forward pass (27.3\,s, 18.1\% of step time) remain on the critical path as sequential stages within each training step, accounting for the bulk of the performance difference.

\FloatBarrier
\subsection{Colocate vs.\ Fully Async}
\label{subsec:exp-modes}

Relax's unified staleness-control abstraction lets users navigate the throughput--freshness tradeoff within a single system.
This section compares colocate mode (training and inference share the same GPUs) with async modes (training and inference on separate nodes) on two configurations that differ in model scale and modality.

\begin{figure*}[htbp]
  \centering
  \begin{subfigure}[b]{0.48\textwidth}
    \centering
    \includegraphics[width=\textwidth]{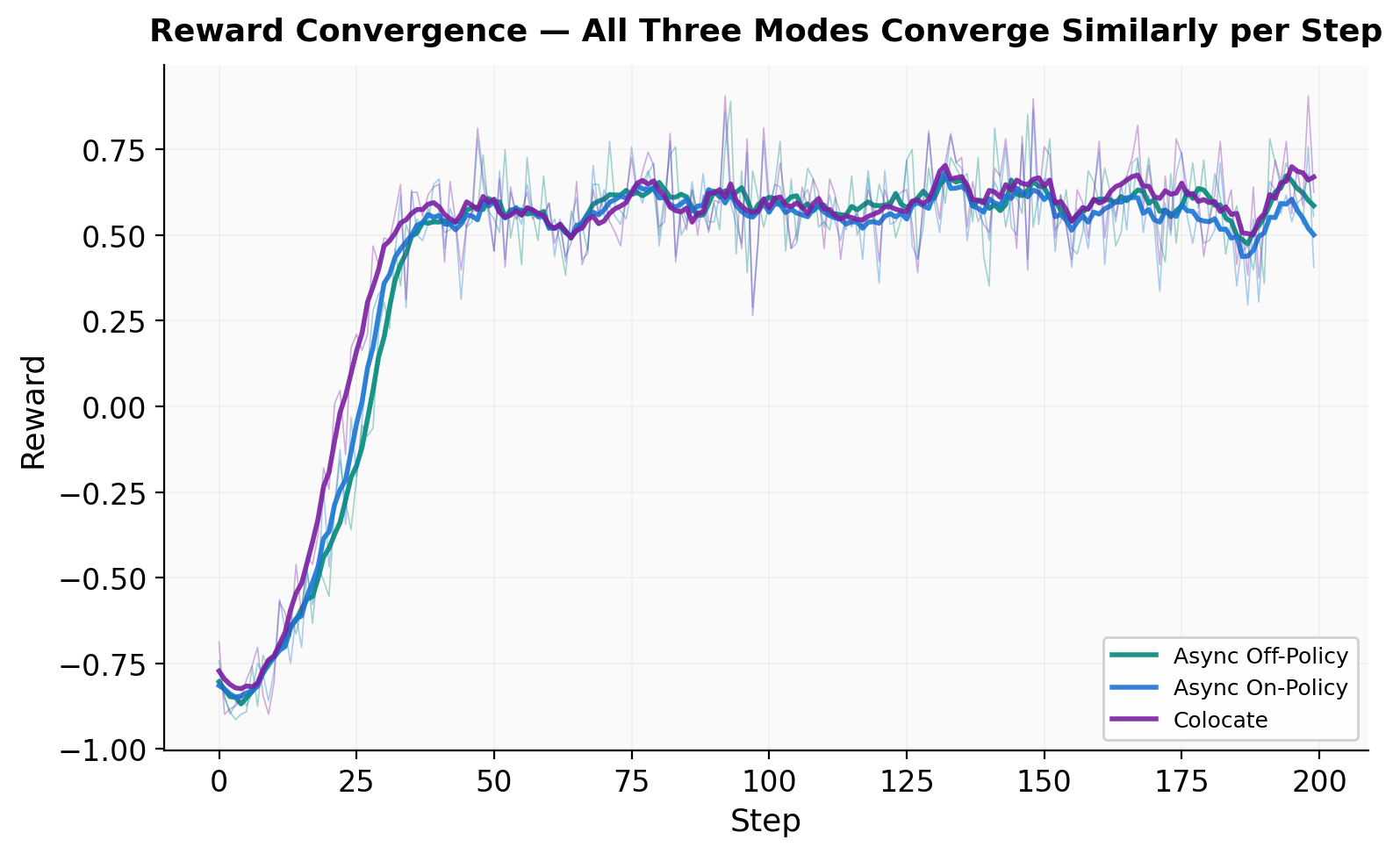}
    \caption{}
    \label{fig:modes-reward-step}
  \end{subfigure}
  \hfill
  \begin{subfigure}[b]{0.48\textwidth}
    \centering
    \includegraphics[width=\textwidth]{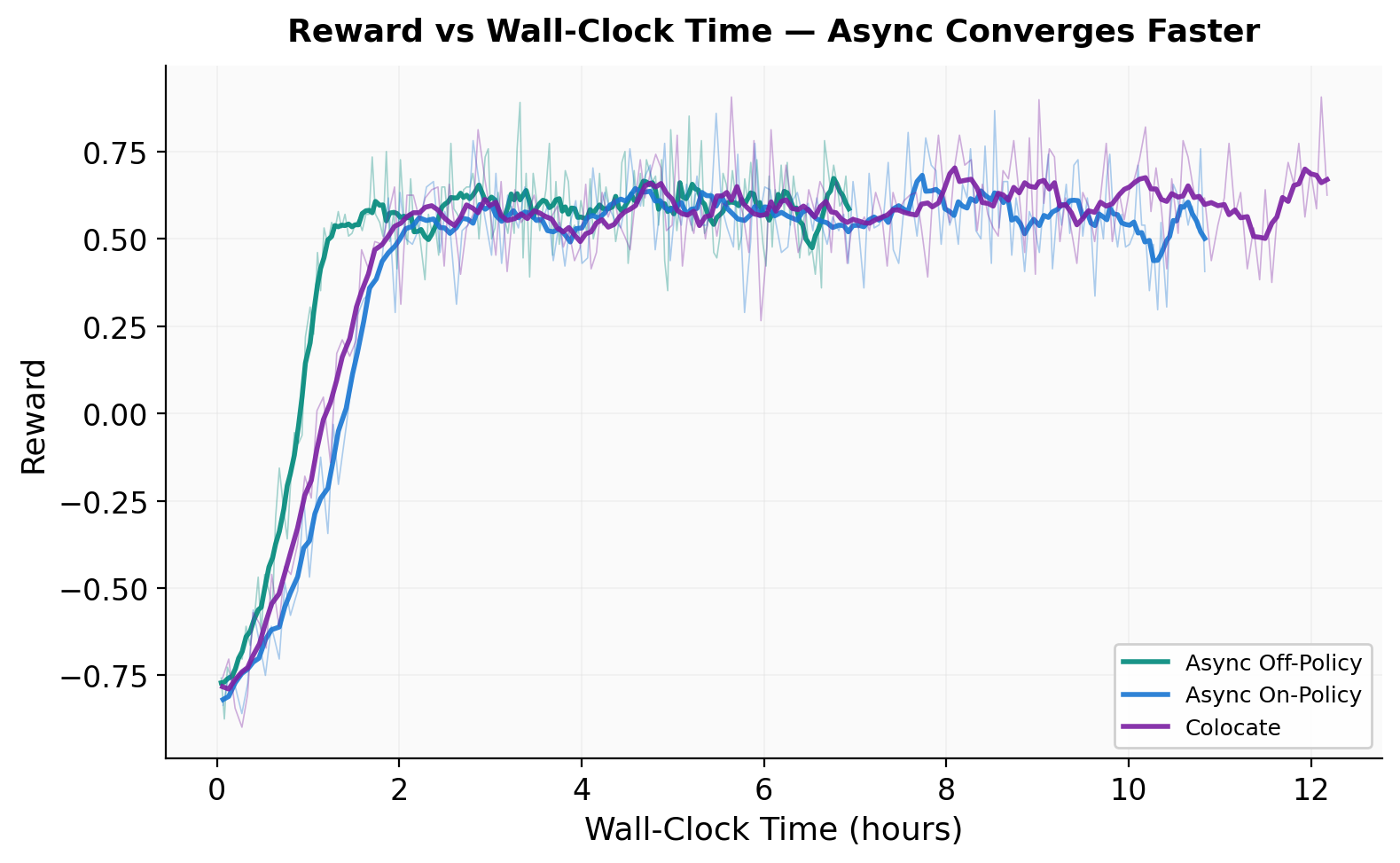}
    \caption{}
    \label{fig:modes-reward-wallclock}
  \end{subfigure}
  \\[6pt]
  \begin{subfigure}[b]{0.48\textwidth}
    \centering
    \includegraphics[width=\textwidth]{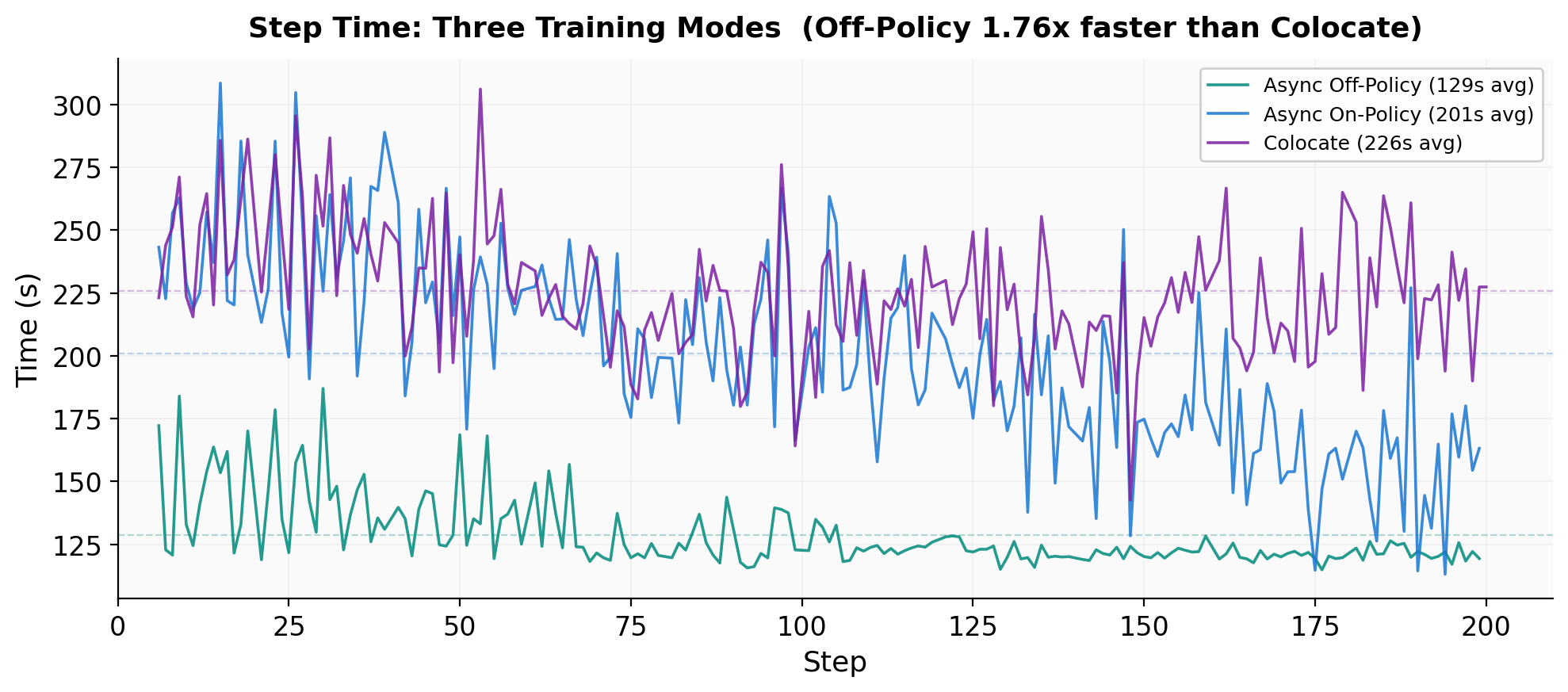}
    \caption{}
    \label{fig:modes-steptime}
  \end{subfigure}
  \hfill
  \begin{subfigure}[b]{0.48\textwidth}
    \centering
    \includegraphics[width=\textwidth]{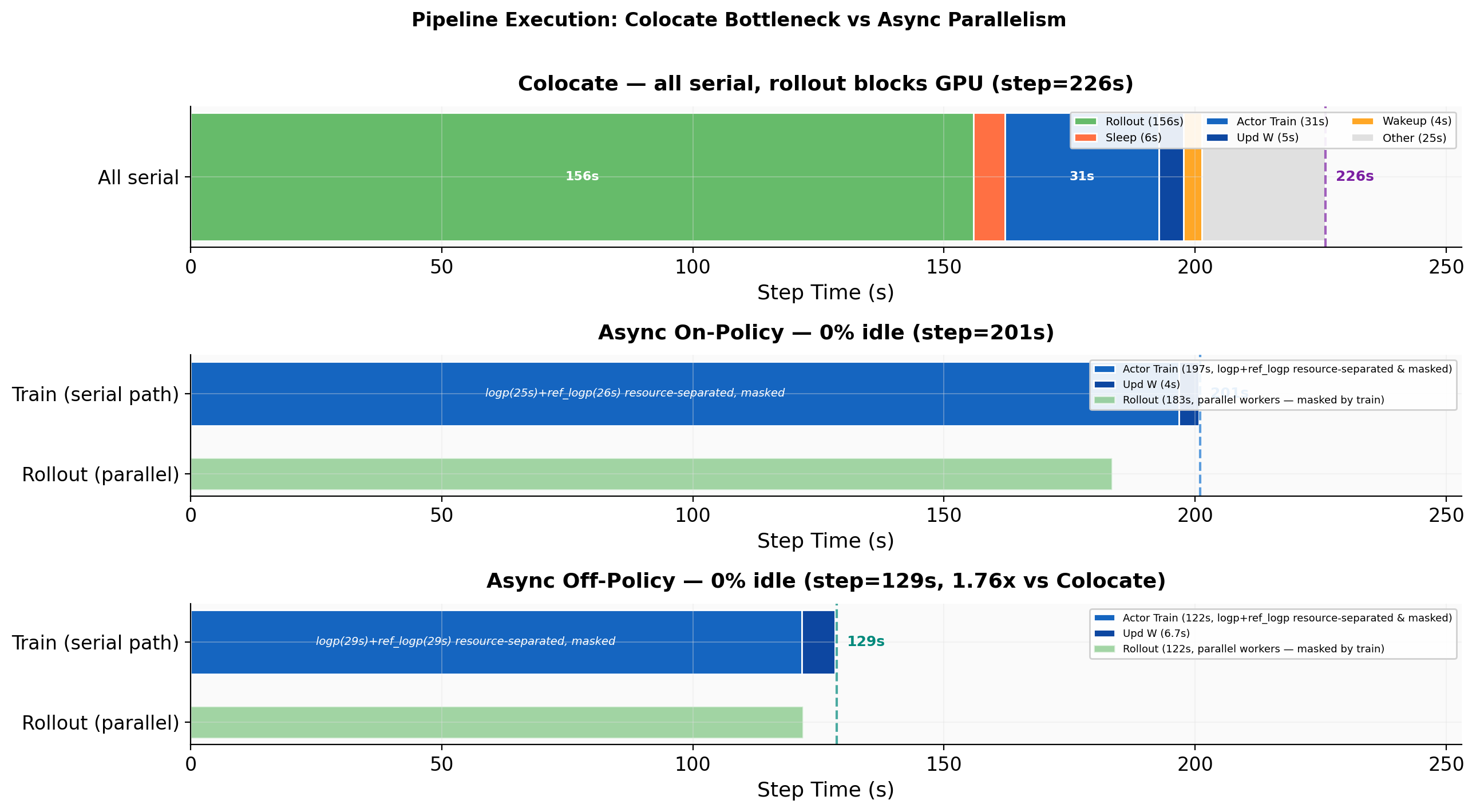}
    \caption{}
    \label{fig:modes-waterfall}
  \end{subfigure}
  \caption{Training-mode comparison on Qwen3-4B / DAPO-MATH-17k. (a)~Reward by step. (b)~Reward by wall-clock time. (c)~Per-step time distribution. (d)~Gantt-chart view of pipeline structure.}
  \label{fig:training-modes}
\end{figure*}

\textbf{Text-only (Qwen3-4B).}\quad
Using Qwen3-4B / 16$\times$H800 / DAPO-MATH-17k, we compare three training modes: off-policy (staleness$=$1), on-policy (staleness$=$0), and colocate.
Async off-policy achieves the highest throughput at 128.6\,s/step (28.0 steps/hr), a \textbf{1.76$\times$ speedup} over colocate (Table~\ref{tab:three-modes}, Figure~\ref{fig:training-modes}).
Async on-policy reaches 201.0\,s/step (17.9 steps/hr), a 1.12$\times$ speedup---demonstrating that even under strict on-policy freshness (staleness$=$0), moving rollout to separate nodes for parallel execution delivers significant gains.
Colocate is the slowest at 225.9\,s/step (15.9 steps/hr), because training and inference must alternate serially on shared GPUs, incurring ${\sim}$10\,s/step sleep/wakeup overhead.
All three modes converge to the same reward level (Figure~\ref{fig:modes-reward-step}), confirming that staleness does not degrade training quality on the DAPO-MATH task; when viewed by wall-clock time, async off-policy reaches the same final reward in roughly half the wall-clock budget (Figure~\ref{fig:modes-reward-wallclock}).

\begin{table}[htbp]
\centering
\caption{Training-mode comparison on Qwen3-4B / DAPO-MATH-17k.}
\label{tab:three-modes}
\footnotesize
\begin{tabular}{@{}llll@{}}
\toprule
\textbf{Metric} & \textbf{Colocate} & \textbf{Async On-Policy} & \textbf{Async Off-Policy} \\
\midrule
Step Time              & 225.9\,s     & 201.0\,s     & \textbf{128.6\,s} \\
Steps/Hour             & 15.9         & 17.9         & \textbf{28.0} \\
vs Colocate Speedup    & 1.00$\times$ & 1.12$\times$ & \textbf{1.76$\times$} \\
Sleep/Wakeup           & 10\,s/step   & 0\,s         & 0\,s \\
\bottomrule
\end{tabular}
\end{table}

\begin{figure*}[htbp]
  \centering
  \begin{subfigure}[b]{0.48\textwidth}
    \centering
    \includegraphics[width=\textwidth]{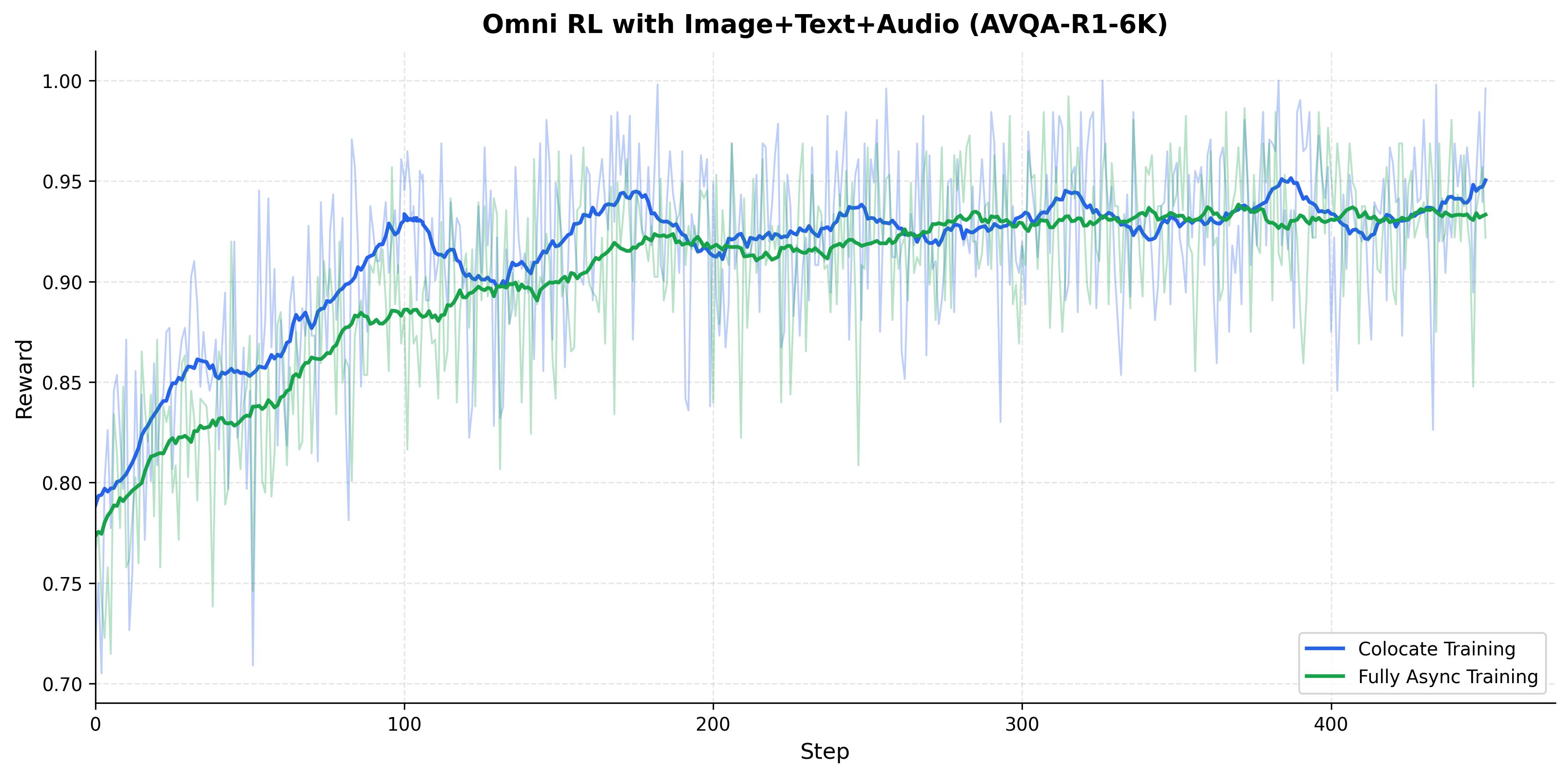}
    \caption{}
    \label{fig:omni-modes-reward-step}
  \end{subfigure}
  \hfill
  \begin{subfigure}[b]{0.48\textwidth}
    \centering
    \includegraphics[width=\textwidth]{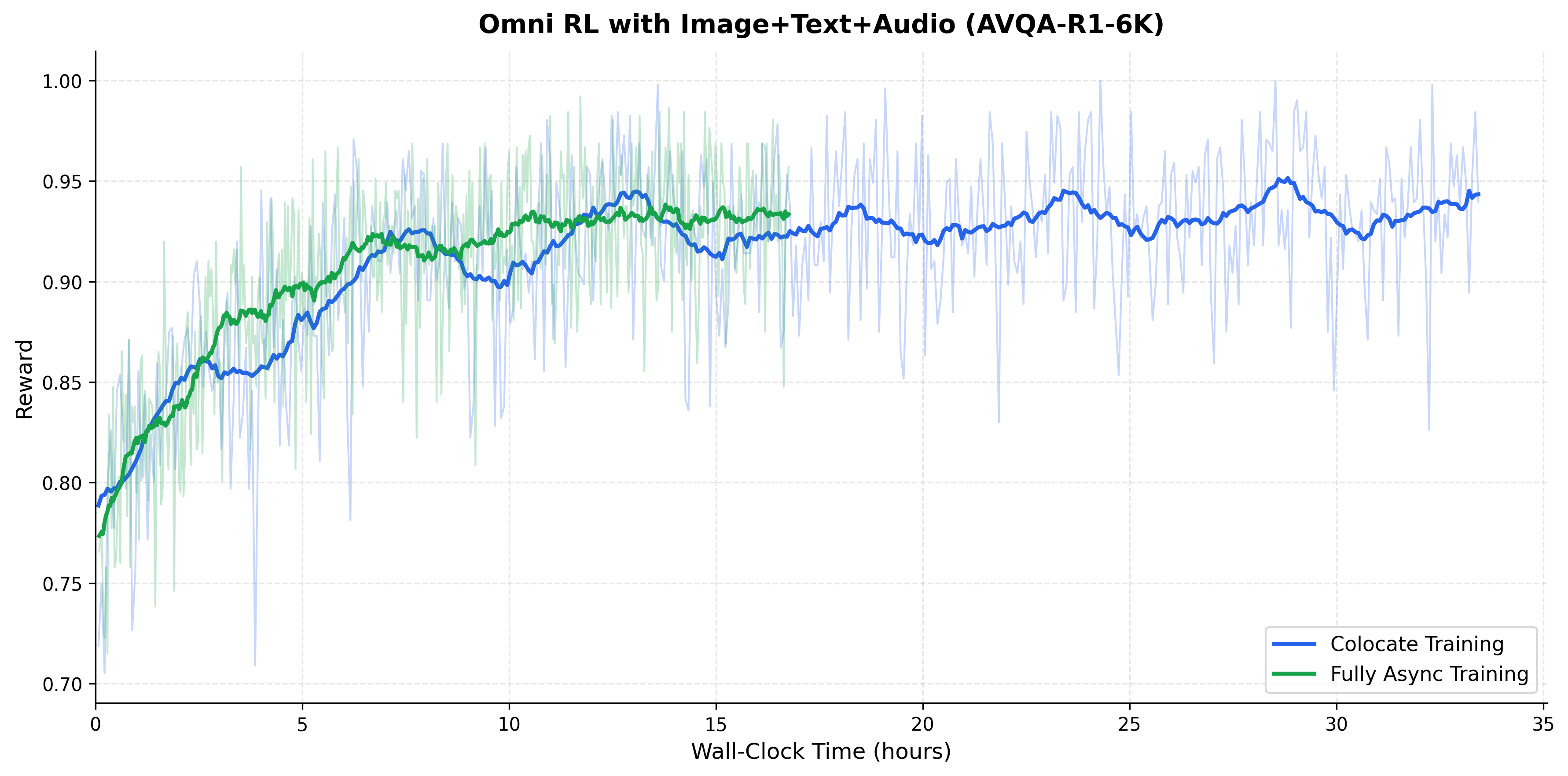}
    \caption{}
    \label{fig:omni-modes-reward-wallclock}
  \end{subfigure}
  \caption{Colocate vs.\ fully async on Qwen3-Omni-30B / Echo Ink. (a)~Reward by step. (b)~Reward by wall-clock time.}
  \label{fig:omni-modes}
\end{figure*}

\textbf{Omni-modal (Qwen3-Omni-30B).}\quad
To verify that the colocate-vs-async tradeoff extends to larger omni-modal models, we compare the same two modes on Qwen3-Omni-30B trained on Echo Ink (audio+image, AVQA-R1-6K) using 16$\times$H20 GPUs, with the async experiment running in off-policy mode (staleness$=$2).
Fully async achieves a \textbf{2.00$\times$ speedup} (133.6\,s vs.\ 267.4\,s per step; 26.9 vs.\ 13.5 steps/hr), and both modes converge to the same final reward of ${\sim}$0.93 (Figure~\ref{fig:omni-modes}a), with the convergence gap narrowing to $<$0.003 at plateau.
The wall-clock view (Figure~\ref{fig:omni-modes}b) makes the throughput advantage concrete: both modes train for the same number of steps, so the fully async curve covers only half the time axis and reaches the ${\sim}$0.93 plateau in roughly half the wall-clock budget.
Compared with the 1.76$\times$ speedup on the 4B text-only model, the larger 2.00$\times$ speedup on the 30B omni-modal model confirms that the async architecture advantage amplifies with model scale---colocate's sleep/wakeup overhead grows proportionally with parameter count (${\sim}$10\,s/step at 4B, already noticeably larger at 30B), while Relax's separated deployment eliminates it entirely.

\FloatBarrier
\subsection{Training Stability: Near-Zero-Overhead R3}
\label{subsec:exp-stability}

MoE models are particularly susceptible to train--inference mismatch in RL training.
Because the inference engine (SGLang) and training framework (Megatron) inevitably produce slightly different numerics, the discrete routing mechanism can flip top-K expert selections for the same token, sending it down an entirely different computation path.
R3 (Rollout Routing Replay)~\cite{ma2025r3} addresses this by recording the top-K expert indices at each MoE layer for every token during inference, then replaying them during training to ensure both engines activate the same experts.

\begin{figure*}[htbp]
  \centering
  \includegraphics[width=\textwidth]{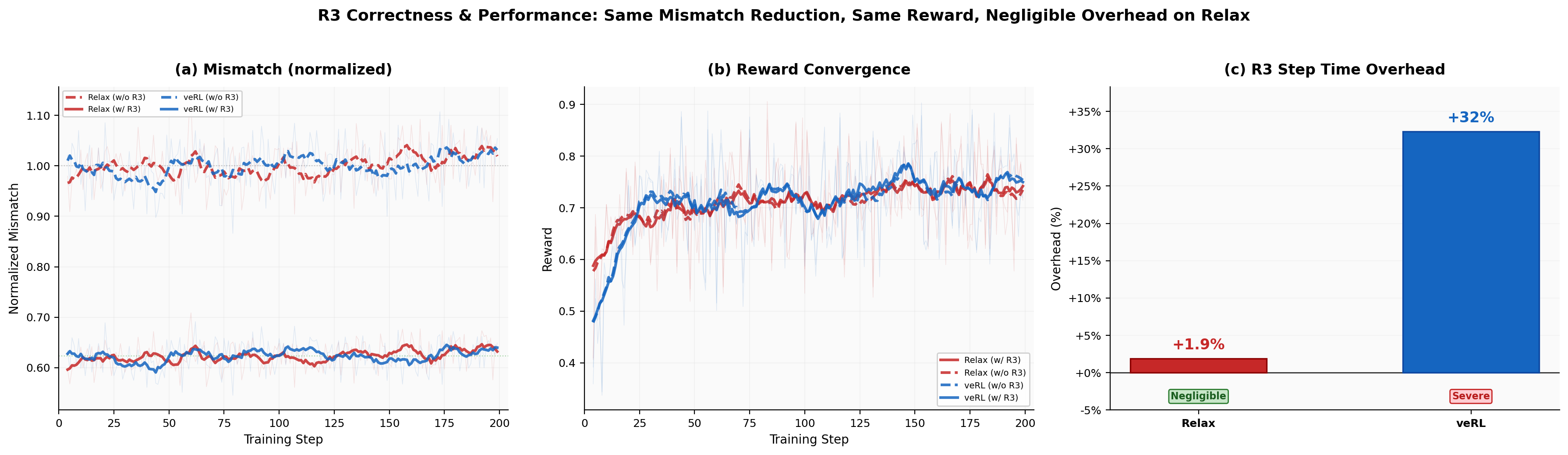}
  \caption{R3 ablation on Qwen3-30B-A3B: Relax vs.\ veRL. (a)~Normalized routing mismatch. (b)~Reward convergence. (c)~Step time overhead (w/ R3 / w/o R3).}
  \label{fig:r3-combined}
\end{figure*}

We evaluate on Qwen3-30B-A3B / 16$\times$H800 / DAPO-MATH-17k, comparing four configurations in a 2$\times$2 matrix (Relax/veRL $\times$ w/wo R3).
R3 reduces rollout--training log-probability mismatch by ${\sim}$38\% on both systems, with the normalized mismatch curves overlapping precisely after R3 is enabled (Figure~\ref{fig:r3-combined}a), and all four configurations follow the same reward trajectory (Figure~\ref{fig:r3-combined}b), confirming that R3 preserves convergence.
Enabling R3 on veRL incurs a \textbf{32\%} step time overhead, whereas Relax absorbs R3 at only \textbf{+1.9\%} (Figure~\ref{fig:r3-combined}c), making R3 a practically free correctness enhancement.
This is achieved by rewriting the serialization path---routing data is extracted from the pickle channel and broadcast via NCCL zero-copy---and keeping broadcast results GPU-resident to eliminate redundant GPU$\to$CPU$\to$GPU round-trips, moving R3 data transfer entirely off the critical path.
Relax additionally supports FP16 training to reduce BF16-induced numerical mismatch; details are in Appendix~\ref{app:fp16}.

\section{Conclusion}
\label{sec:conclusion}

We presented Relax, an open-source RL training engine that co-designs three tightly coupled dimensions---omni-modal data handling, service-isolated deployment, and staleness-unified asynchronous training---to address the challenges of next-generation RL workloads.
The central insight is that these dimensions form a causal chain: omni-modal data introduces workload heterogeneity that demands service-level fault isolation; service decoupling enables an asynchronous data bus that reconciles throughput with policy freshness; and the field-based data model of the bus naturally accommodates heterogeneous modality fields.

Experiments validate this co-design across multiple axes.
On end-to-end throughput, Relax achieves a 1.20$\times$ speedup over veRL on Qwen3-4B on-policy training.
On training-mode flexibility, its fully async off-policy mode delivers a 1.76$\times$ speedup over colocate on Qwen3-4B and a 2.00$\times$ speedup on Qwen3-Omni-30B, while all modes converge to the same reward level.
On training stability, enabling R3 routing replay on veRL incurs a 32\% step time overhead, whereas Relax absorbs R3 at only +1.9\%, making it a practically free correctness enhancement for MoE models.
On convergence breadth, we demonstrate stable RL training on Qwen3-Omni-30B across image, text, audio, and video modalities over 2{,}000 steps, and verify correct multi-turn agentic RL on tool-calling tasks.

Relax is open-sourced at \url{https://github.com/redai-infra/Relax}.
We hope it provides a useful foundation for teams working on omni-modal and agentic RL training at scale.

\section*{Acknowledgments}
We thank the Slime team at Zhipu AI, the NVIDIA Megatron Bridge team, and the Ascend TransferQueue team for their open-source contributions that Relax's infrastructure builds upon.

\bibliographystyle{unsrt}
\bibliography{references}

\newpage
\appendix
\section{Discussion}
\label{sec:discussion}

\subsection{Design Trade-offs}
\label{subsec:tradeoffs}

\paragraph{The role of colocate mode.}
Although the fully asynchronous (off-policy) mode delivers the highest throughput in our experiments, colocate mode remains a practical choice when the GPU budget is limited.
By time-sharing GPUs between training and inference, colocate mode achieves reasonable throughput with roughly half the total GPU count compared to separate-cluster deployments.
The 78.1\% idle ratio reported in the training mode comparison (\S\ref{subsec:exp-perf}) reflects GPU time-sharing between the training and inference phases, not wasted computation---the actual compute stages (rollout + training $\approx$ 50.95\,s) occupy a small fraction of the step.
For teams with constrained hardware budgets, this tradeoff is often acceptable.

\paragraph{Service-oriented complexity cost.}
The service-oriented architecture introduces deployment complexity compared to monolithic training scripts.
Initial setup requires configuring Ray Serve, the Distributed Checkpoint Service (DCS), and TransferQueue---components that must be deployed and monitored as independent services.
This overhead is justified for long-running (multi-day) training campaigns where fault tolerance and debuggability save more time than they cost: a single unrecoverable failure in a monolithic system can waste hours of GPU time, whereas Relax's two-tier recovery (\S\ref{sec:server-arch}) restores individual roles in minutes.
However, for short experimental runs (e.g., a few hundred steps for hyperparameter exploration), the added setup overhead may not be warranted, and a simpler monolithic script may be more practical.

\paragraph{TransferQueue: benefits vs.\ overhead.}
TransferQueue interposes a storage layer between data producers and consumers, adding serialization and network transfer overhead.
This cost is amortized when training and inference run on separate GPU clusters (the fully asynchronous case), where the network transfer latency would exist regardless of the data-bus abstraction.
For colocate deployments with shared memory, the overhead is measurable but modest relative to the flexibility gained---in particular, the ability to switch between training modes by changing a single staleness parameter without code modifications.

\paragraph{Reproducibility considerations.}
Our convergence experiments use proprietary datasets (Echo Ink, Deepeyes), which limits external reproducibility of the convergence results specifically.
We mitigate this limitation in three ways: (1)~throughput experiments use publicly described configurations (Qwen3-4B on DAPO-MATH-17k) that can be independently replicated; (2)~the text CoT experiments on Qwen3-30B-A3B produce reward curves aligned with veRL's published results on the same model and dataset, providing cross-validation; and (3)~the Relax framework itself is fully open-sourced, enabling the community to verify system-level claims on their own datasets and models.

\subsection{Limitations and Future Work}
\label{subsec:limitations}

While Relax provides comprehensive support for omni-modal \emph{understanding} tasks (image, text, audio, and video perception followed by language-based reasoning), it does not yet support RL training for \textbf{generative modalities}---i.e., training policies that produce images, video, or audio as outputs.
Extending the pipeline to generative RL (e.g., text-to-image reward optimization) requires new reward interfaces, modality-specific decoding strategies, and evaluation metrics, and is an active area of future development.

Our convergence experiments span models up to 35B parameters (dense) and 30B-A3B (MoE).
\textbf{Ultra-large-scale} validation on models with 397B+ parameters is ongoing; preliminary results are promising but not yet mature enough for inclusion in this report.

While the service-oriented architecture introduces significant operational benefits, it also adds \textbf{deployment complexity} relative to monolithic training scripts.
Organizations with limited infrastructure expertise may find the initial setup more involved, though we believe this cost is amortized over the improved debuggability and fault tolerance during long-running training campaigns.

\paragraph{Elastic scaling.}
Relax's service-oriented architecture enables dynamic adjustment of rollout replicas during training via an HTTP REST API, without restarting the job.
Two scaling modes are supported: \emph{Ray-native} (intra-cluster) scaling adds or removes replicas within the current Ray cluster, while \emph{external} (cross-cluster) scaling allows replicas from separate clusters to join via TCP-based weight synchronization.
Both modes share a unified replica lifecycle state machine (\emph{pending} $\to$ \emph{initializing} $\to$ \emph{ready} for scale-out; \emph{draining} $\to$ \emph{removed} for scale-in), ensuring weight consistency throughout transitions.
An optional Autoscaler service triggers scaling decisions based on GPU utilization, queue depth, or time-to-first-token latency.
This capability is implemented but not yet included in the current open-source release; we plan to open-source it in a future version.

\paragraph{Richer agentic scenarios.}
Ongoing work includes SWE-Agent (software engineering tasks with code execution), CodeAgent (interactive debugging), and search-augmented reasoning, all of which benefit from Relax's multi-turn and tool-use support.


\section{Supplementary Experiments}
\label{app:supplementary}

\subsection{Text-Only RL Convergence on DAPO-MATH}
\label{app:dapo-math}

While \S\ref{subsec:exp-convergence} focuses on omni-modal convergence, text-only RL remains an important baseline for verifying system correctness.
We train Qwen3-30B-A3B with DAPO on DAPO-MATH-17k using 16$\times$H800 GPUs.

\begin{figure*}[htbp]
  \centering
  \begin{subfigure}[b]{0.48\textwidth}
    \centering
    \includegraphics[width=\textwidth]{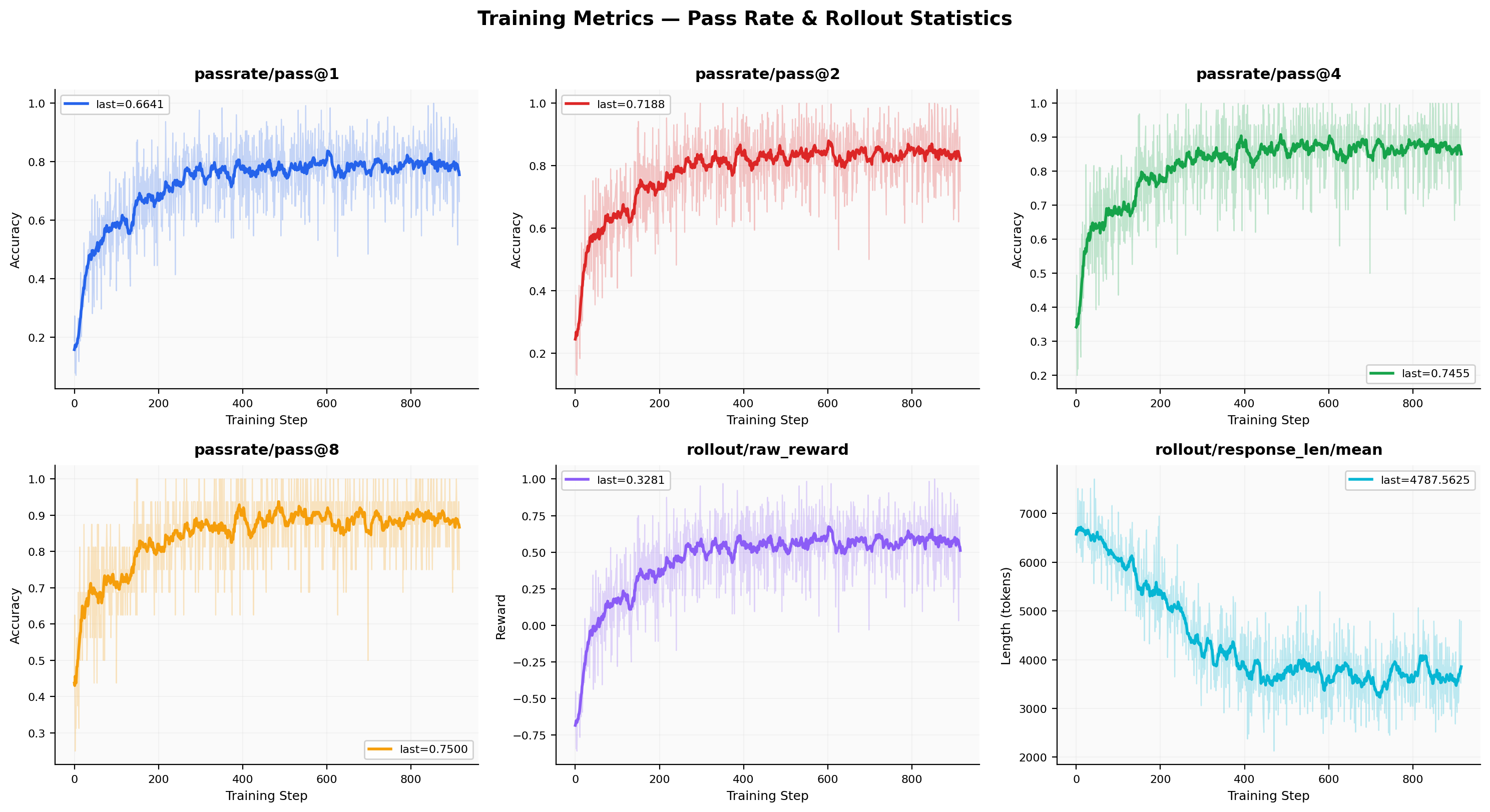}
    \caption{Training metrics: reward, pass@k, and response length.}
    \label{fig:dapo-train}
  \end{subfigure}
  \hfill
  \begin{subfigure}[b]{0.48\textwidth}
    \centering
    \includegraphics[width=\textwidth]{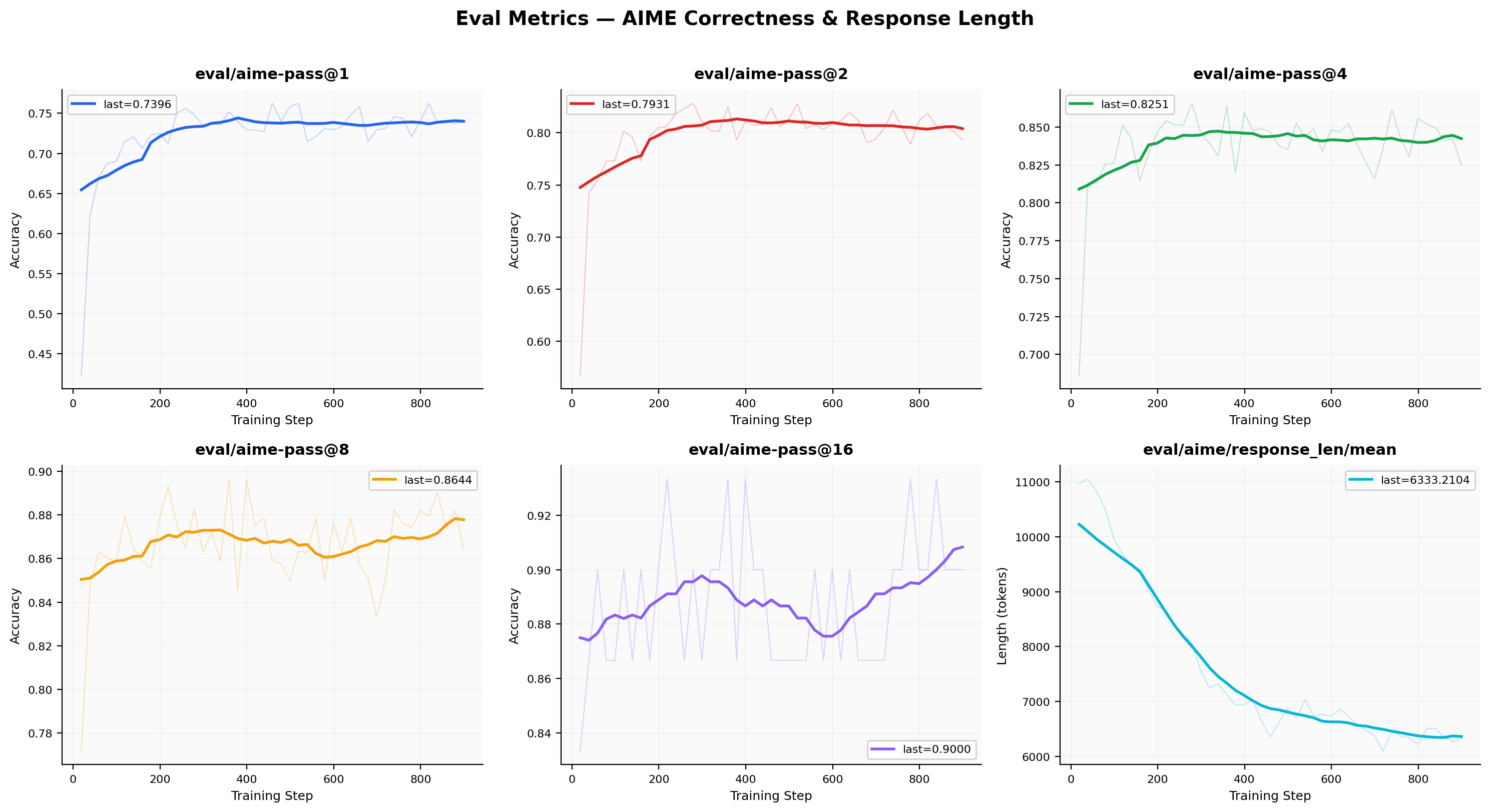}
    \caption{AIME-24 evaluation pass@k over training steps.}
    \label{fig:dapo-eval}
  \end{subfigure}
  \caption{Text-only RL on DAPO-MATH (Qwen3-30B-A3B). (a)~Training reward, pass@k, and response length. (b)~AIME-24 evaluation pass@k over training steps.}
  \label{fig:dapo-math}
\end{figure*}

Training reward and pass@k rates improve steadily (Figure~\ref{fig:dapo-train}), with response length increasing as the model learns longer chain-of-thought reasoning.
AIME-24 evaluation scores rise monotonically throughout training (Figure~\ref{fig:dapo-eval}), confirming that training gains transfer to held-out benchmarks.
These results verify that Relax's service-isolated architecture and TransferQueue data bus introduce no correctness regressions on standard text RL workloads.

\subsection{Multimodal Agentic RL Convergence}
\label{app:agentic-rl}

\begin{figure}[htbp]
  \centering
  \includegraphics[width=\columnwidth]{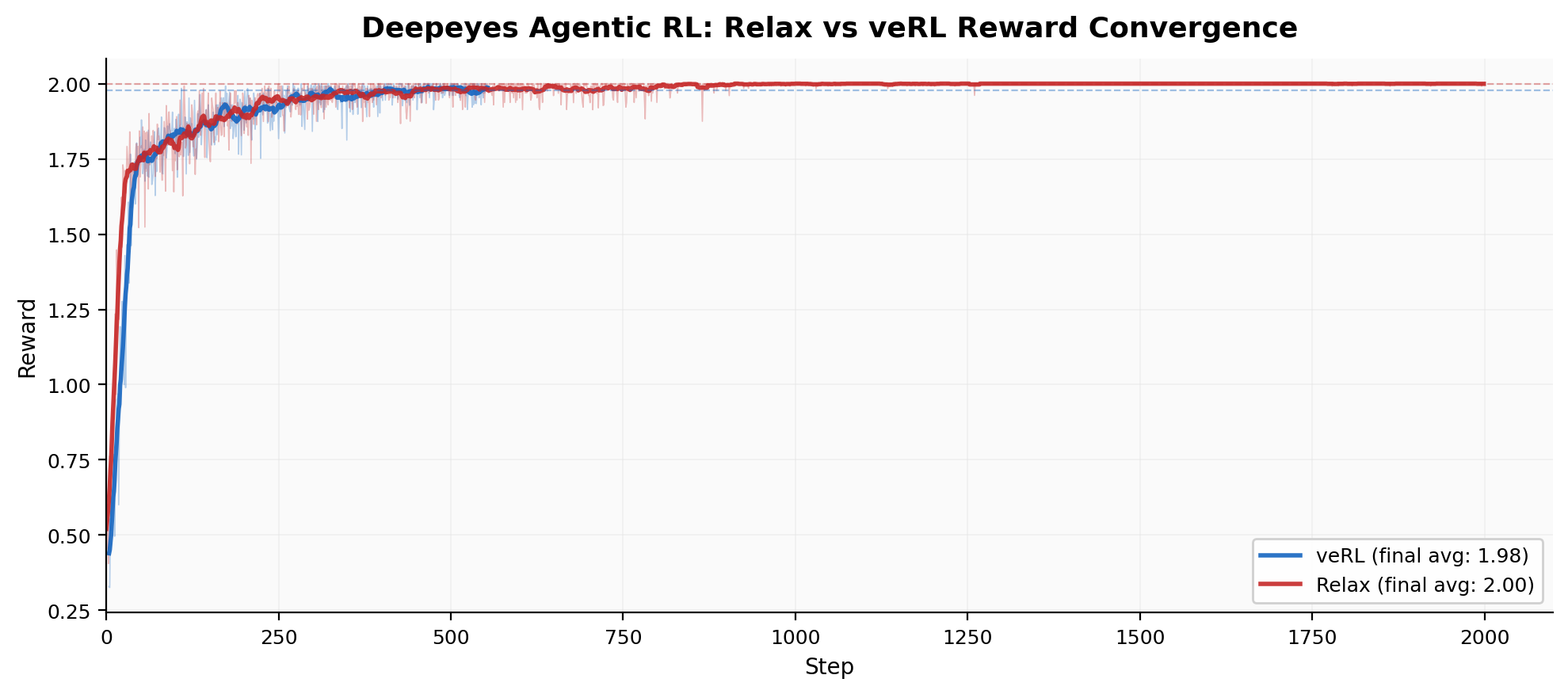}
  \caption{Agentic RL reward convergence on multi-turn tool-calling (Qwen3-VL-MoE-30B, Deepeyes).}
  \label{fig:conv-deepeyes}
\end{figure}

Agentic RL introduces a distinct correctness challenge: variable-length, multi-turn trajectories where each turn's observation depends on the previous action, requiring correct reward propagation and advantage estimation across turns.
We evaluate on the Deepeyes visual QA task, where the model executes multi-step tool-calling sequences (image crop, zoom) to answer image-based questions.
Both Relax and veRL train Qwen3-VL-MoE-30B in colocate mode under the same configuration.

Both systems exhibit nearly identical S-curve convergence trajectories, reaching the reward upper bound of 2.0 (Figure~\ref{fig:conv-deepeyes}), with final 50-step averages of 2.0000 (Relax) and 1.9783 (veRL).
This confirms that Relax's trajectory collection, advantage estimation, and policy gradient computation remain correct under multi-turn agentic workloads, validating the extension points described in \S\ref{sec:agentic}.

\subsection{FP16 vs.\ BF16 Precision Ablation}
\label{app:fp16}

As noted in \S\ref{subsec:exp-stability}, R3 addresses routing mismatch for MoE models.
For dense models, a complementary source of train--rollout divergence is floating-point precision: BF16's limited mantissa (7 bits) amplifies rounding divergence between the Megatron training path and the SGLang inference path, increasing log-probability mismatch.
FP16's 10-bit mantissa is more robust to these path differences, yielding tighter consistency.

We evaluate FP16 vs.\ BF16 on Qwen3-4B using 8$\times$H800 GPUs in colocate mode with GRPO.

\begin{figure*}[htbp]
  \centering
  \begin{subfigure}[b]{0.48\textwidth}
    \centering
    \includegraphics[width=\textwidth]{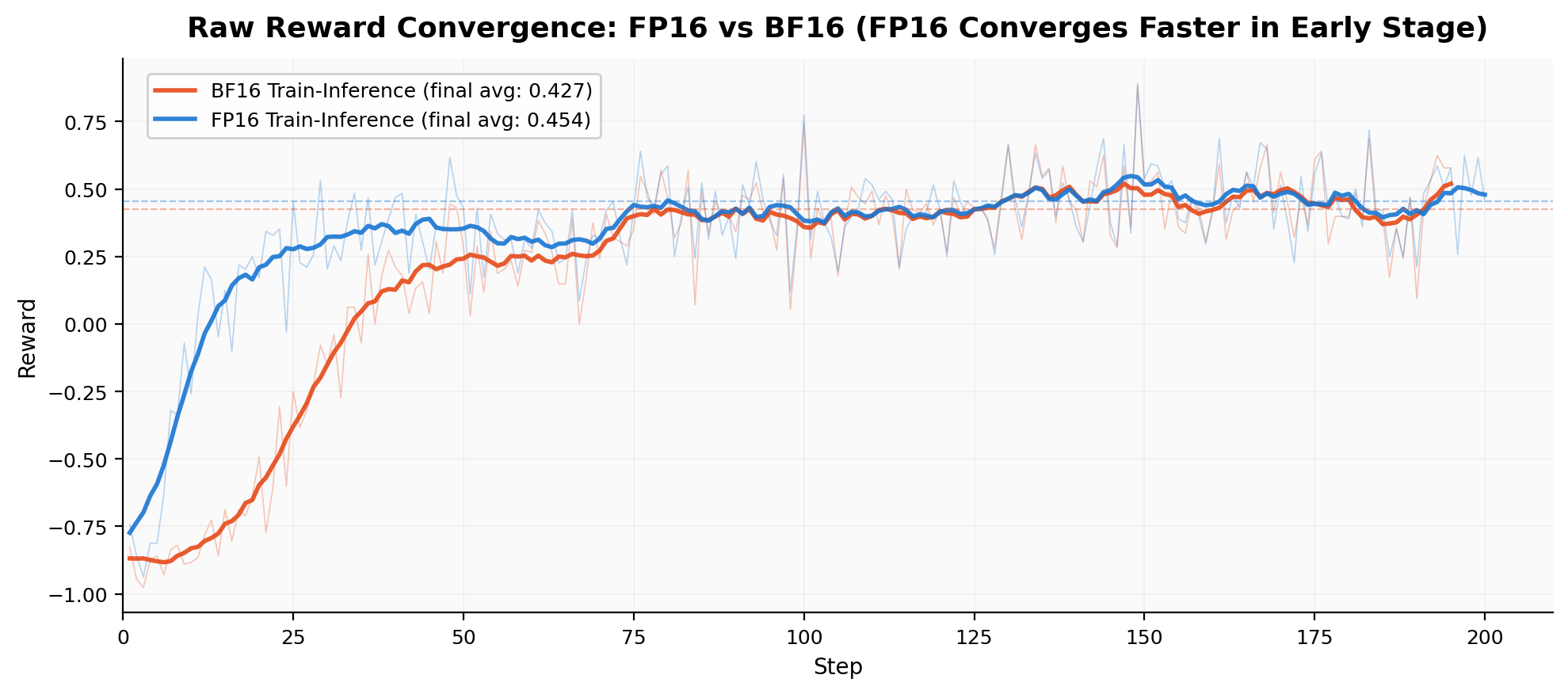}
    \caption{Reward convergence.}
    \label{fig:fp16-reward}
  \end{subfigure}
  \hfill
  \begin{subfigure}[b]{0.48\textwidth}
    \centering
    \includegraphics[width=\textwidth]{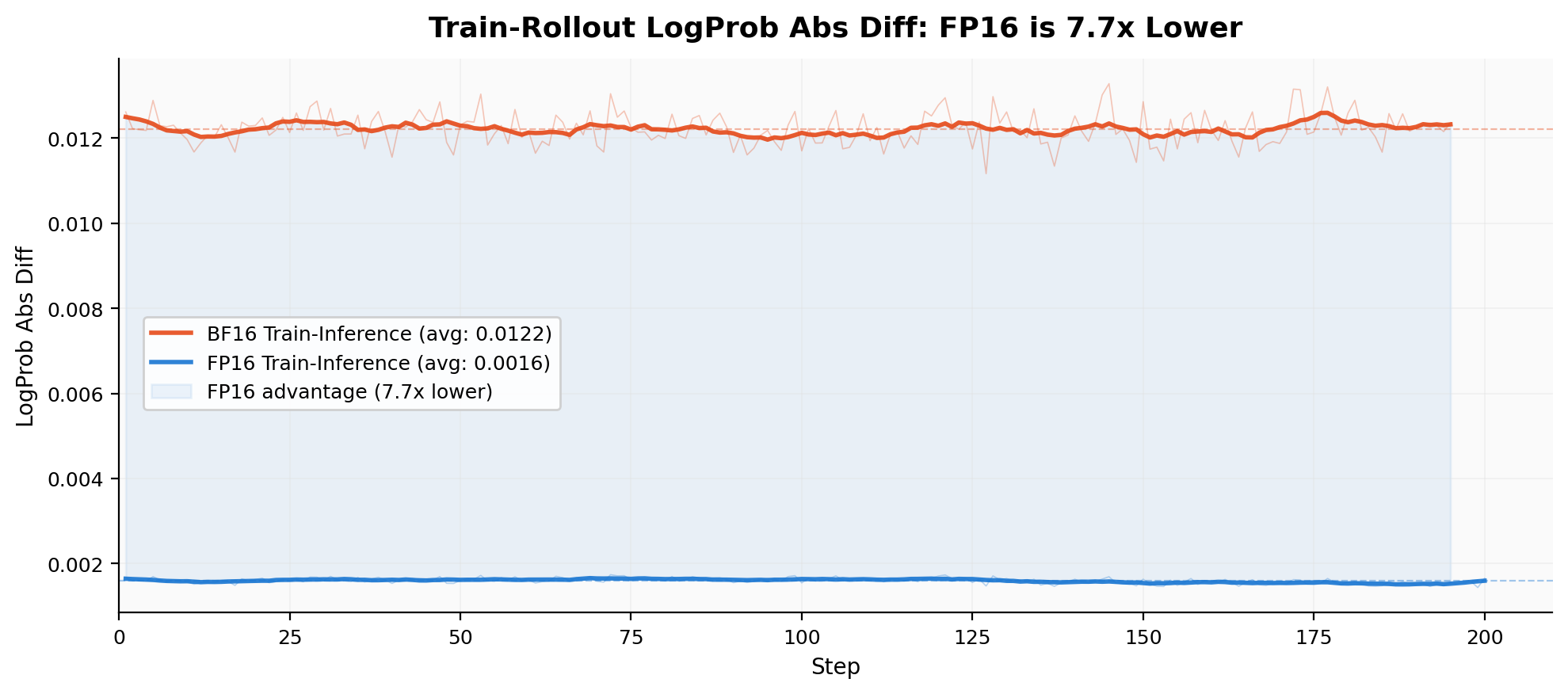}
    \caption{Train--rollout log-probability mismatch.}
    \label{fig:fp16-logprob}
  \end{subfigure}
  \caption{FP16 vs.\ BF16 precision ablation (Qwen3-4B). (a)~Reward convergence. (b)~Train--rollout log-probability absolute difference.}
  \label{fig:fp16-ablation}
\end{figure*}

Under FP16, the train--rollout log-probability absolute difference drops from 0.0122 (BF16) to 0.0016---a 7.7$\times$ reduction (Figure~\ref{fig:fp16-logprob}).
This lower mismatch yields more accurate importance sampling ratios in GRPO, explaining the faster early-stage convergence: FP16 enters positive reward territory at approximately 15 steps, compared to 30 steps for BF16 (Figure~\ref{fig:fp16-reward}).
Both configurations converge to comparable final reward levels (difference $<$0.03), and FP16's log-probability difference remains stable throughout training with near-zero variance, confirming numerical reliability.
Together with R3 for MoE models, FP16 support gives practitioners two complementary levers for minimizing train--rollout numerical divergence.

\end{document}